\documentclass{article} 
\usepackage{nips13submit_e,times}
\usepackage{url}
\usepackage{mahnigCEC}
\usepackage{epsfig}
\usepackage{hyperref}

\title{Gray-box optimization and factorized distribution algorithms: where two worlds collide}

\author{
Roberto Santana \\
Intelligent Systems Group\\
University of the Basque Country (UPV/EHU)\\
\texttt{roberto.santana@ehu.es} 
}

\nipsfinalcopy 

\begin{document}

\maketitle

\begin{abstract}
  
  The concept of gray-box optimization, in juxtaposition to black-box optimization, revolves about the idea of exploiting the problem structure to implement more efficient evolutionary algorithms (EAs). Work on  factorized distribution algorithms (FDAs), whose factorizations are directly derived from the problem structure, has also contributed to show how exploiting the problem structure produces important gains in the efficiency of EAs. In this paper we analyze the general question of using problem structure in EAs focusing on confronting work done in gray-box optimization with related research accomplished in FDAs. This contrasted analysis helps us to identify, in current studies on the use problem structure in EAs, two distinct analytical characterizations of how these algorithms work. Moreover, we claim that these two characterizations collide and compete at the time of providing a coherent framework to investigate this type of algorithms. To illustrate this claim, we present a contrasted analysis of formalisms, questions, and results produced in FDAs and gray-box optimization. Common underlying principles in the two approaches, which are usually overlooked, are identified and discussed.  Besides, an extensive review of previous research  related to different uses of the problem structure in EAs is presented. The paper also elaborates on some of the questions that arise when extending the use of problem structure in EAs, such as the question of evolvability, high cardinality of the variables and large definition sets, constrained and multi-objective problems, etc. Finally, emergent approaches that exploit neural models to capture the problem structure are covered.   


\end{abstract}

{\bf{keywords}}:  genetic algorithms, problem structure, gray-box optimization, structure modeling, estimation of distribution algorithms, probabilistic graphical models, neural models, evolvability



\section{Introduction}

A number of recent works in evolutionary algorithms have emphatically highlighted the need of exploiting the problem structure information when addressing optimization problems with evolutionary algorithms (EAs) \cite{Whitley:2015a,Whitley2016a}. Expressions such as ``problem structure matters'' \cite{Whitley:2015a}  or ``blind (search) no more'' \cite{Whitley2016a} refer to the excruciating importance of identifying and exploiting problem information. Furthermore, the term \emph{gray-boy optimization} has been coined to refer to a variety of algorithms, all sharing the characteristic of using,  to different extents,  the problem structure information. Research in this direction fosters the idea that exploiting  information about the problem characteristics can produce important gains in efficiency.  Gray-box optimization research then attempts new ways to solve the optimization problems for which some a-priori information is available, notably those problems with a  ``suitable'' structure. The underlying assumption is that knowing this type of ``structural'' information can not only serve to improve traditional EA implementations, but also to create significantly novel and more efficient approaches.


 Presenting ways of using knowledge about the problem characteristics in the design of genetic algorithms (GAs) \cite{Goldberg:1989,Holland:1975} was an early topic of attention in EAs \cite{Beasley_et_al:1993,Clearwater_and_Hogg:1994,Maini_et_al:1994}.  DeJong \cite{DeJong:1988} mentions two main approaches to adapt the classical GA definition to the characteristics of the problem: 1) To design an alternative representation of the same (solution) space for which the traditional (genetic) operators are appropriate. 2) To select different genetic operators that are more appropriate to the ``natural representation''. These two approaches have been present in several GA applications, typically in those that implement knowledge-based or heuristic genetic operators  \cite{Butz_et_al:2006,Goldberg_et_al:1993,Riff:1997,TianLi_et_al:2009}.

 For the analysis made in this paper we will assume that the ``problem structure'' exploited by gray-box optimizers corresponds to a singular representation of the a-priori known characteristics of the problem. Furthermore, we will assume that what makes gray-box optimization depart from other EA approaches, such as those mentioned above, is the conjunction of a particular type of problem knowledge (the structure of the  interactions among variables) with very specific methods to exploit this knowledge. Therefore,  while  knowledge-based operators have been extensively investigated in evolutionary computation, we will examine the methods conceived for exploiting the problem structure, such as those advocated by gray-box optimization, as novel.  

 
 Problem structure, which is used by gray-box optimizers, has been previously exploited by other EAs. First and foremost, by model-building EAs, which create a representation of the relationships among the variables of the problem and use this representation to conduct a more efficient sampling of the search space. These algorithms have undergone rapid development in recent years \cite{Bosman_and_Grahl:2008,Grahl_et_al:2008,Larranaga_et_al:2012,Shakya_and_Santana:2012a,Zhou_et_al:2008,Zlochin_et_al:2004}.   Significant research has been conducted in the field of estimation of distribution algorithms (EDAs) \cite{Larranaga_et_al:2012,Muhlenbein_and_Paas:1996r}  to better exploit information about the problem structure while solving the optimization problem. 

With the exception of the simplest EDAs that assume independence between variables (e.g., PBIL \cite{Baluja:1994} and UMDA \cite{Muhlenbein:1997}), the  best known variant of EDAs are those that  automatically learn probabilistic  models of (black-box) optimization problems during the search. Nevertheless, there are EDAs whose factorizations are directly derived from the problem structure. They were originally called factorization-based distribution algorithms (FDAs) \cite{Muhlenbein_et_al:1999}. The analysis presented in this paper uses  the similarities and differences between FDAs and gray-box optimization methods as a leitmotif to discuss several points involved in using problem information by EAs. 



 The goal of the paper is three-fold: First, to identify those concepts that are essentially identical in the two approaches, bridging the gap between analyses originated from different perspectives of the same problem. Secondly, to review research on the exploitation of the graphical representation of the problem structure in EAs. Finally, our aim is to discuss relevant questions related to the use of the problem structure in EAs, including the emergence of neural models,  approaches that exploit the structure of multi-objective and constrained problems, and  directions for convergence between gray-box optimization and EDAs. 
 

 The paper is organized as follows: The ideas brought up by works on gray-box optimization and gray-box optimizers  are discussed in the next section.  Some background on EDAs, and particularly on FDAs, is presented in Section~\ref{sec:EDA_BG}.  The relationship between gray-box optimization and EDAs is analyzed from different perspectives in Section~\ref{sec:GRAY_EDA}. Section~\ref{sec:GRAPHS} illustrates the application of hyperplane-based and factorization-based formalisms to the analysis of a decomposable function.  Section~\ref{sec:FUTURE} discusses a number of relevant topics and challenging questions for EAs that exploit the problem structure. Section~\ref{sec:CONCLU} concludes the paper.

\section{Gray-box optimization problems and gray-box optimizers} \label{sec:GRAYBOX_BG}

 To introduce the ideas of gray-box optimization, we briefly address the following questions:

  \begin{enumerate}
   \item A definition of problem structure.
   \item An introduction to black-box, gray-box, and a finer color scale for problem structure characterization in optimization.
   \item An explanation of what gray-box optimization is about.
  \end{enumerate}

 \subsection{Problem structure}

 We will assume in this paper that \emph{problem structure} refers to the specific patterns of interactions among variables at the time of determining the objective function. In loose terms, a problem with no structure is a problem where all variables influence independently the values of the objective function. On the contrary, a problem with difficult or intricate structure is one in which several distinct large subsets of interacting variables, that partially overlap, participate in the computation of the function values.

 The notion of problem structure we use here is related but essentially different to the one implicit in the analysis of fitness landscapes \cite{Gallagher_and_Yuan:2006,Hernando_et_al:2017,Kerschke_et_al:2015,Mitchell_et_al:1992a}, where the focus is not necessarily on the patterns of interactions among the variables but on other  properties of the problem (e.g., number of local optima, basins of attraction, geometrical landscape features, etc.). Although we will mainly focus on single-optimization problems, the notion of problem structure can be naturally extended to multi-objective problems for which we can still evaluate how different patterns of interactions among the variables influence each of the objectives  \cite{Aguirre_and_Tanaka:2007,Karshenas_et_al:2014,Lopez_et_al:2014,Okabe_et_al:2004,Zangari_et_al:2016}.


 \subsection{Gray-box optimization problems}

      Gray-box optimization problems are  usually explained in contraposition to a black-box optimization problems. In the latter, we do not have any information about the structure of the function being optimized. In the first, case however, we know some information about the structure of the problem. For instance, in additively decomposed functions (ADFs) \cite{Muhlenbein_et_al:1999},  the value of the function is the  sum of the evaluation of a number of subfunctions defined on subsets of all the variables. If we consider that these subsets of (interacting) variables define the structure of the problem, and this structure is known, then ADFs can be seen as a gray-box optimization problem. Several real-world problems could be included in the class of  gray-box, e.g., MAX-kSAT, Ising model, NK-landscape, etc. \cite{Whitley:2015}. 

While gray-box optimization is a relatively recent concept, the term  \emph{gray-box identification} \cite{Pearson_and_Pottman:2000} is given a similar meaning in modeling physical and networked systems and in control theory.  It refers to situations in which a generic model structure is given and the parameters are those to be estimated from data.

  
Even if the difference between black-box and gray-box optimization problems seems sufficiently clear, there are situations in which information about the problem exists but it is only partial. For instance, we could know which are the groups of related variables where subfunctions are defined, but not the way in which they are related (i.e., the expression for the subfunctions defined in each group).  It is also possible that structural relationships are only known for a limited number of groups, i.e., some definition sets of the function are unknown.

 For combinatorial problems, we introduce in this paper a finer grain definition of the type of available problem information and the way it characterizes the  optimization problem. We split the available information into: 1) Information about the structural relationships among variables (definition sets). 2) Information about the way in which the interactions are expressed within each group (definition of the subfunctions). Table~\ref{tab:WHITE_GRAY_BLACK} shows the White-Gray-Black (WGB) classification of optimization problems according to the type and extent of problem information available. Using this classification, black-box optimization problems would be classified as Black-Black problems, and gray-box optimization problems could be further divided into another $6$ groups. To avoid confusion, and for ease of presentation, we will stick to the ``gray-box'' term in this paper.

 \begin{table*}
\begin{center}
\scriptsize
\begin{tabular}{c|c|l}
\hline\hline
   Structure  & Subfunction def.  &  Type of problems \\ \hline \hline  
   White      & White     &  All definition sets and their corresponding subfunctions are known. \\
   White      & Gray      &  All definition sets are known but some subfunctions are unknown. \\ 
   White      & Black      & All definition sets are known but no information about subfunctions is available. \\ 
   Gray       & White      & Definition sets are partially known, together with all their corresponding subfunctions. \\ 
   Gray       & Gray       & Definition sets are partially known and only some of their corresponding subfunctions are available. \\ 
   Gray       & Black      & Definition sets are partially known. No information about the subfunctions is available.  \\ 
   Black      & Black      &  Nothing is known about the structure and consequently the subfunctions.\\  \hline \hline  
\end{tabular}
\caption{The White-Gray-Black (WGB) classification of optimization problems according to the type and extent of problem information available. The information is classified using two criteria:  1) Definition sets of the function (Function structure). 2) Expression or procedure to define each subfunction (Subfunction definition). \emph{White} refers to the case where the information is fully available. \emph{Gray} to the situation in which the information is partially known, and \emph{Black} when there is no available information.}
\label{tab:WHITE_GRAY_BLACK}
\end{center}
\end{table*}

\subsection{Gray-box optimizers}
 
 Gray-box optimizers  are optimization algorithms that exploit the information available about the structure of an optimization problem in order to make a more efficient search.  It is assumed that, at least for some classes of gray-box problems, using this information will produce gains in efficiency.  However, this is a very general definition since different classes of optimizers can use information about the problem structure in distinct ways and with different degrees of efficiency. 

  In the papers where the term gray-box optimization is discussed  \cite{Chicano_et_2014,Chicano_et_al:2016,Tinos_et_al:2015,Whitley_et_al:2013},  it is usually assumed that information about the structure is known \emph{a priori}, i.e., it is not learned by the algorithm itself while conducting the search. Furthermore,  while there are no apparent reasons to set constraints on the structural characteristics  in the general class of gray-box optimization problems, for feasibility and efficiency reasons, gray-box optimizers assume that the  structure of the problem is constrained. For instance, it is assumed that the size of any definition subset of an ADF to be optimized is upper-bounded by a parameter $k$ (e.g.,  \emph{k-bounded pseudo-Boolean} functions as presented in \cite{Whitley:2015}).

  A number of methods have been covered under the umbrella of gray-box optimizers. From highly efficient hill-climbers \cite{Chicano_et_al:2016,Whitley_et_al:2013}, to enhanced partition crossover operators \cite{Chicano_et_2014,Tinos_et_al:2015},  and combinations of black-box global optimizers and local-search gray-box optimizers \cite{Goldman_et_al:2015}.

\section{Estimation of distribution algorithms} \label{sec:EDA_BG}
  
The main idea of Estimation of distribution algorithms (EDAs) \cite{Larranaga_et_al:2012,Lozano_et_al:2005,Muhlenbein_and_Paas:1996r}  is to extract patterns shared by the best solutions, represent these patterns using a probabilistic graphical model (PGM) \cite{Koller_and_Friedman:2009,Pearl:2000}, and generate new solutions from this model. In contrast to GAs, EDAs apply learning and sampling of distributions instead of classical crossover and mutation operators. Modeling the dependencies among the variables of the problem serves to efficiently orient the search to more promising areas of the search space by explicitly capturing and exploiting potential relationships among the problem variables.  The pseudocode of an EDA is shown in Algorithm~\ref{alg:EDA}.

\small
\begin{BAlgo}{Estimation of distribution algorithm}
	\label{alg:EDA}
	\item Set $t\Leftarrow 0$. Create a population $D_0$ by generating $N$ random solutions.
	\item \Do
	\item \T {Evaluate $D_t$ using the fitness function.}
	\item \T {From $D_t$, select a population $D_t^S$ of $K \leq N$ solutions according to a selection method.}
	\item \T {Compute a probabilistic model of $D_t^S$.}
	\item \T {Generate $D_{t+1}$ sampling from the model.}
	\item \T {$t \Leftarrow t+1$}
	\item  \Until{Termination criteria are met.}
\end{BAlgo}
\normalsize

\subsection{Factorized distribution algorithms} 

 While it is usually assumed that EDAs learn the structure of the problem from data, this is not always the case. In fact, the first EDA based on the theory of PGMs was called Factorized Distribution Algorithm (FDA) \cite{Muhlenbein_et_al:1999} and it computed a factorization from a problem structure known  apriori. In the field of EDAs, the term FDA was initially used to refer to EDAs that  learned only the parameters of the  probabilistic models and not the structure  \cite{Muhlenbein_and_Mahnig:1998,Muhlenbein_and_Mahnig:1999a,Muhlenbein_and_Mahnig:1999,Santana_et_al:1999,Santana_et_al:2001c,Zhang:2004}. However, the term was later also used to cover EDAs that learn the structure from data \cite{Mahnig:2001b,Muehlenbein_and_Hoens:2006,Muehlenbein_and_Mahnig:2002,Ochoa_et_al:2000a,Ochoa_et_al:1999,Santana:2003c,Santana_et_al:2001b}. In this sense, the ``FDA'' and ``EDA'' terms were both used indistinctly to refer to the same class of algorithm. While the term ``EDA'' attempts to emphasize the role played in the algorithm by the \emph{distribution estimation step}, the term ``FDA'' highlights that a \emph{factorization of the distribution}  is used in the estimation.  

 FDAs that use a fixed, a priori known structure of the problem to factorize the distribution were mainly applied for only a short time due to the rapid introduction of EDAs capable to learn higher order factorizations directly from the data, such as those based on Bayesian networks \cite{Etxeberria_and_Larranhaga:1999,Larranhaga_et_al:1999,Pelikan_et_al:1999,Soto_et_al:1999} and Markov networks \cite{Alden_and_Miikkulainen:2016,Santana:2003c,Shakya_and_McCall:2007}. This may explain why FDAs based on a fixed structure are known to only a reduced number of early EDA adopters. However, and this is one of the main claims made in this paper, several of the questions originally addressed for the first FDAs (those based on a fixed structure), and the answers given to these questions, are related and are a deep concern to current research in gray-box optimization.

\subsection{Alternative views of EDAs}

 In order to understand the different implications of using probabilistic modeling in EAs, and the significance of taking into consideration the problem structure in the construction of the models,   it is convenient to approach EDAs from different perspectives. In this section we briefly discuss four of these perspectives. 

 \subsubsection{An EA that learns and samples a probability model}

   The most common definition of an EDA is an EA which replaces crossover and mutation operators by the process of learning and sampling a model. This understanding of EDAs emphasizes their difference to GAs. 

 \subsubsection{An automatic way to generate models of an optimization problem} 

  When structural learning of the probabilistic model is applied, EDAs can be seen as methods that iteratively capture the ``hidden'' structure of the optimization problem. This means, in each generation, the learning method ``mines'' the selected population as traditional data-mining methods \cite{Hastie_et_al:2001} do and unearths a model of the relations in the population.

In some applications, the model learned can be as advantageous as finding the optimal solution. Such a type of structure can serve to define similarity relationships among different instances of the same problem \cite{Hauschild_et_al:2008,Santana_et_al:2013}, or to design transfer learning strategies \cite{Kaedi_and_Ghasem:2011,Pelikan_and_Hauschild:2012,Santana_et_al:2012f}. In this sense EDAs can be seen as  an automatic way to generate models of the problem. Furthermore, since these models are produced in a temporarily ordered manner, it is possible to  associate the  structural characteristics of the models to different stages of the search.

\subsubsection{An effective method for problem decomposition} \label{sec:PROBDECOMP}

 Problem decomposition is a general goal in Artificial Intelligence. The implicit parallelism hypothesis used to explain GAs assumes that multiple subproblems are simultaneously solved during the GA evolution, eventually leading to the optimal solution. A fundamental question for the design of effective GAs is to find a tight encoding of the building blocks so to avoid their disruption, the so-called linkage problem \cite{Harik_and_Goldberg:1996,Holland:1975}. FDAs with a fixed model solve this problem by appropriately encapsulating and respecting the interactions during the sampling process, effectively exchanging the building blocks of the parents in the new population.

 EDAs that learn the structure actually identify the building blocks of the problem and organize them in the model  in such a way that they will be usable for sampling. Both the process of identifying those building blocks (learning), and the process of combining them by sampling, are two essential ingredients of the automatic method for problem decomposition as implemented in EDAs. The problem decomposition perspective is applicable for methods, not necessarily EDAs, that can reuse the building blocks for implementing other alternative ways for improving the solutions \cite{Mendiburu_et_al:2012,Santana_et_al:2013b,Sastry_and_Goldberg:2004a}. This perspective of problem decomposition is also closely linked to gray-box optimization, showing a path for inclusion of gray-box optimization algorithms in EAs \cite{Goldman_et_al:2015}. 

\subsubsection{An algorithm to evolve probability distributions}

An EDA can be seen as a process in which the different components of the algorithm assign a probability to every solution of the search space. To illustrate this fact, we introduce some basic notation.

Let $ {\bf{X}}=(X_1,\ldots ,X_n)$ be a vector of discrete variables. We will  use ${\bf{x}}=(x_1,\ldots ,x_n)$ to denote an assignment to the variables. $S$ will denote a set of indices in $\{1, \ldots, n\}$, and $X_S$ (respectively $x_S$) a subset of the variables of ${\bf{X}}$ (respectively ${\bf{x}}$) determined by the indices in $S$.

\begin{figure}
\begin{center}
\includegraphics[width=14.0cm]{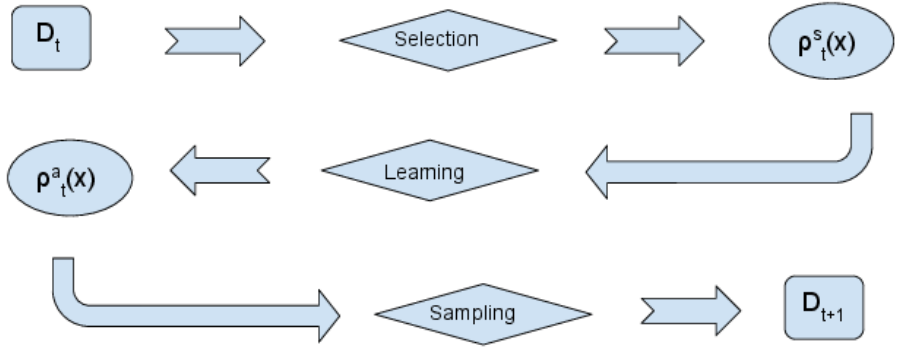}
\caption{Joint probability distributions determined by the components of an   EDA. $D_{t}$, $D_{t+1}$: populations at generation $t$ and $t+1$; $\rho_{t}^s({\bf{x}})$, $\rho_{t}^a({\bf{x}})$: Joint probability distributions determined by selection and the probabilistic model  approximation.}
\label{fig:SELINEDAS}
\end{center}
\end{figure}

Let  $D_{t}$ and $D_{t+1}$ denote the EDA populations at generations $t$ and $t+1$, respectively. The selection method defines a probability of selection ($\rho_{t}^s({\bf{x}})$) for each solution ${\bf{x}}$. Similarly, the factorization encoded by the probabilistic model defines a probability distribution that assigns a probability $\rho_{t}^a({\bf{x}})$ of appearing in the next population to each solution. Figure~\ref{fig:SELINEDAS} shows one possible representation of the way in which EDA components define the probability distributions of the solutions.

 The GA recombination operators could be also seen as defining probability distributions on the solution space. What makes EDAs different is that the probabilistic models make the  probabilities assigned to the solutions explicit. In GAs, approximating the probabilities determined by crossover and mutation can be cumbersome. In EDAs, given a probabilistic model, the computation of the probability is straightforward. Looking at EDAs from the perspective of algorithms that move in the space of probabilities is pertinent for the theoretical analysis of the algorithms \cite{Echegoyen_et_al:2012a,Echegoyen_et_al:2013,Lozada_and_Santana:2011,Muehlenbein_and_Mahnig:2001b,Muehlenbein_and_Mahnig:2002a} and can serve as a basis for implementing different types of inferences about the characteristics of the search space.

\section{Gray-box optimization and EDAs} \label{sec:GRAY_EDA}

 In this section we analyze different aspects of  the relationship between  gray-box optimization and EDAs. In particular, the following aspects will serve to illustrate these relationships:

 \begin{enumerate}
   \item Class of problems used to study the algorithms.
   \item Representation of the problem structure.
   \item Modeling the search for optimal solutions.
   \item Using the problem structure for the search. 
   \item Implications of the structure for the behavior of the algorithms. 
 \end{enumerate}

\subsection{Class of problems}


In additively decomposable functions (ADFs), possible interactions among the variables are reduced to a subset of these interactions. ADFs are used to simulate problems that can be decomposed into smaller subproblems. 
 
\begin{definition} 
  An ADF of order $k$ is  defined by 
 
 \begin{equation} 
   f(x)=\sum_{s_i\in S}f_i(\Pi _{s_i}{\bf{x}})\qquad S=\{s_1,\dots ,s_l\}\quad 
   s_i\subseteq X, |s_i| = k 
\end{equation} 
\end{definition} 
where $S$ is the set comprising the definition sets of the function, and $\Pi _{s_i}{\bf{x}}$ is the projection of ${\bf{x}} \in {\bf{X}}$ onto the subspace $X_S$. 

Other approaches in EAs conceptualize the notion of ADF with different names. For instance, an embedded landscape with bounded epistasis $k$ is defined in  \cite{Heckendorn:2002}  as a function that can be decomposed  as the sum of $m$ subfunctions, each one depending at most on $k$ input variables.

ADFs have been extensively used to investigate the influence of the  problem structure and study the convergence properties of EDAs \cite{Brownlee_et_al:2012a,Echegoyen_et_al:2011,Muhlenbein_and_Mahnig:1999,Pelikan_and_Hauschild:2012a,Zhang:2004}. The behavior of these algorithms for other decomposable problems has been also studied from the perspective of ADFs \cite{Grahl_et_al:2008a,Sastry_et_al:2005a}. Linkage identification methods have been analyzed using ADFs \cite{Chen_et_al:2012,Chuang_and_Chen:2007,Rohlfshagen_and_Bullinaria:2008,Tsuji_et_al:2007,Zhou_et_al:2008}. 

When $x_i \in \{0,1\}$,  $k$-order ADFs are essentially identical to \emph{k-bounded pseudo-Boolean}, functions which are used in several works addressing gray box optimization \cite{Chicano_et_2014,Whitley:2015,Whitley_et_al:2016}. For instance, in  \cite{Whitley:2015},  any problem that is expressed as a k-bounded pseudo-Boolean optimization problem, and for which $M=O(n)$  is called an \emph{Mk landscape}. 
In this paper, we stick to the terms of $k$-order ADFs and $k$-order decomposable problems since these terms are self-explanatory.

\subsection{Representation of the problem structure}

  One of the key points in the analysis of the problem structure is to find meaningful representations of the relationships among the variables that allow the exploitation of the potential patterns hidden in the structure. In this section we analyze the graphical representations of the structure usually applied in gray-box optimization and EDAs, and the way these representations have been exploited. 
 
 \subsubsection{Interaction graphs}

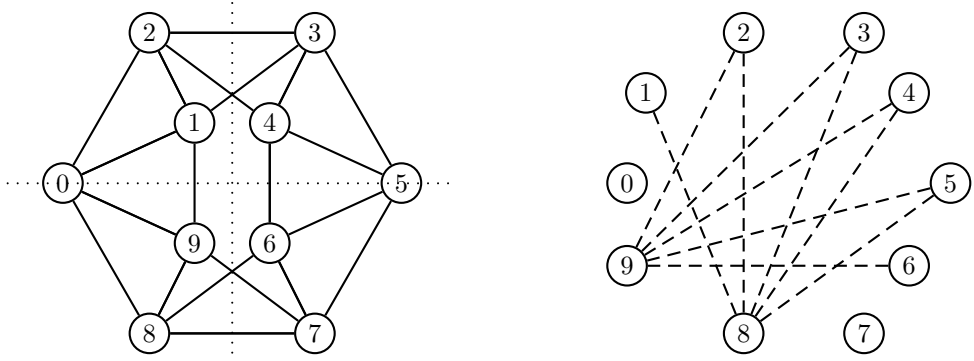
\begin{figure}[htb] 
\begin{center} 
\begin{pspicture}(0,0)(14,5) 

\rput(0.75,2){\circlenode{P0}{$ 0$}} 
\rput(2.5,2.8){\circlenode{P1}{$ 1$}} 
\rput(1.9,4){\circlenode{P2}{$ 2$}}
\rput(4.1,4){\circlenode{P3}{$ 3$}}
\rput(3.5,2.8){\circlenode{P4}{$ 4$}}
\rput(5.25,2){\circlenode{P5}{$ 5$}} 
\rput(3.5,1.2){\circlenode{P6}{$ 6$}} 
\rput(4.1,0){\circlenode{P7}{$ 7$}} 
\rput(1.9,0){\circlenode{P8}{$ 8$}} 
\rput(2.5,1.2){\circlenode{P9}{$ 9$}}

\ncline{-}{P0}{P1} \ncline{-}{P1}{P2}  \ncline{-}{P0}{P2} 
\ncline{-}{P1}{P2} \ncline{-}{P1}{P3}  \ncline{-}{P2}{P3}
\ncline{-}{P2}{P3} \ncline{-}{P2}{P4}  \ncline{-}{P3}{P4} 
\ncline{-}{P3}{P4} \ncline{-}{P4}{P5}  \ncline{-}{P3}{P5}
\ncline{-}{P4}{p5} \ncline{-}{P4}{P6}  \ncline{-}{P4}{P6}
\ncline{-}{P5}{P6} \ncline{-}{P6}{P7}  \ncline{-}{P5}{P7}
\ncline{-}{P6}{P7} \ncline{-}{P7}{P8}  \ncline{-}{P6}{P8}
\ncline{-}{P7}{P8} \ncline{-}{P8}{P9}  \ncline{-}{P7}{P9}
\ncline{-}{P8}{P9} \ncline{-}{P9}{P0}  \ncline{-}{P8}{P0}
\ncline{-}{P9}{P0} \ncline{-}{P0}{P1}  \ncline{-}{P9}{P1}
\psline[linestyle=dotted](3.0,-0.4)(3.0,4.4)
\psline[linestyle=dotted](0,2.0)(6.0,2.0)

\rput(8.25,2){\circlenode{P0}{$ 0$}} 
\rput(8.5,3.2){\circlenode{P1}{$ 1$}} 
\rput(9.8,4){\circlenode{P2}{$ 2$}}
\rput(11.4,4){\circlenode{P3}{$ 3$}}
\rput(12.0,3.2){\circlenode{P4}{$ 4$}}
\rput(12.55,2){\circlenode{P5}{$ 5$}} 
\rput(12.0,0.9){\circlenode{P6}{$ 6$}} 
\rput(11.4,0){\circlenode{P7}{$ 7$}} 
\rput(9.8,0){\circlenode{P8}{$ 8$}} 
\rput(8.25,0.9){\circlenode{P9}{$ 9$}}

\ncline[linestyle=dashed]{-}{P1}{P8}
\ncline[linestyle=dashed]{-}{P2}{P8}
\ncline[linestyle=dashed]{-}{P3}{P8}
\ncline[linestyle=dashed]{-}{P4}{P8}
\ncline[linestyle=dashed]{-}{P5}{P8}
\ncline[linestyle=dashed]{-}{P2}{P9}
\ncline[linestyle=dashed]{-}{P3}{P9}
\ncline[linestyle=dashed]{-}{P4}{P9}
\ncline[linestyle=dashed]{-}{P5}{P9}
\ncline[linestyle=dashed]{-}{P6}{P9}

\end{pspicture} 
\caption{Left) Interaction graph for a problem with $n=10$ variables and $M=10$ subfunctions of order $k=3$. Horizontal and vertical axes are represented with dotted lines to emphasize the symmetries that exist in the problem structure. Right) Edges added to the original interaction  graph to obtain a chordal graph.} 
\label{fig:INTGRAPH} 
\end{center} 
\end{figure}

   In gray-box optimization, a  variable interaction graph (VIG)  \cite{Chicano_et_2014,Whitley_and_Chen:2012} is defined by associating one vertex to each variable of the problem. Every pair of variables that appear together in some subfunction are connected by an edge in the VIG. Figure~\ref{fig:INTGRAPH}a) shows  an interaction graph for a problem with $n=10$ variables and $M=10$ subfunctions of order $k=3$. 

The VIG represented in  Figure~\ref{fig:INTGRAPH}a) clearly displays, as triangles, the three-way interactions among subsets of variables. In addition to the local patterns, the graph seems to indicate a global symmetric pattern with respect to horizontal and vertical axes. Variables symmetries \cite{Choi_et_al:2007,Munteanu_and_Rosa:2007} are only one example of the types of regularities of a problem structure that could be exploited by ``structure-informed'' optimization methods \cite{Picek_et_al:2015,Santana_et_al:2013a}.  

  Usually, EDAs use probabilistic graphical models  (PGMs) to represent a factorization of the probability distribution. The PGM contains a graphical representation of the dependency relationships among the variables. Common PGMs are dependency trees, Bayesian networks, and Markov networks. Although PGMs are the norm when addressing black-box optimization problems for which the structure has to be learned, interaction graphs were early applied in EDAs \cite{Muhlenbein_and_Hoens:2005,Muehlenbein_and_Hoens:2006,Santana_et_al:2005b} to investigate the relationship between the structure of the problem and the factorized approximation used to solve them.  Table~\ref{tab:REL_FDA} shows a cross-index of related concepts in gray-box optimization and EDAs.

   

\begin{table*}
\begin{center}
\scriptsize
\begin{tabular}{l|l|l|l}
\hline\hline
  gray-box  & EDAs & reference EDAs & reference gray-box \\ \hline \hline  
   Mk landscape problem               & k-order ADF            &   \cite{Whitley_and_Chen:2012}              &   \cite{Muhlenbein_et_al:1999}      \\ 
   
 structure-aware mutation             & BB-wise mutation       &  \cite{Chicano_et_2014,Whitley_and_Chen:2012}               &  \cite{Lima_et_al:2006,Mendiburu_et_al:2012,Sastry_and_Goldberg:2004a}        \\ 
variable interaction graph            &interaction graphs      &   \cite{Chicano_et_2014,Whitley_and_Chen:2012}            &    \cite{Muhlenbein_and_Hoens:2005,Muehlenbein_and_Hoens:2006,Santana_et_al:2005b}     \\ 
 Tree decomposition Mk landscapes  & ADF with bounded  tree-width of $k$ &  \cite{Whitley_et_al:2016}   &  \cite{Muhlenbein_et_al:1999}      \\ \hline \hline  
\end{tabular}
\caption{A cross-index of related concepts in gray-box optimization and EDAs.}
\end{center}
\label{tab:REL_FDA}  
\end{table*}

\subsubsection{Factor graphs}

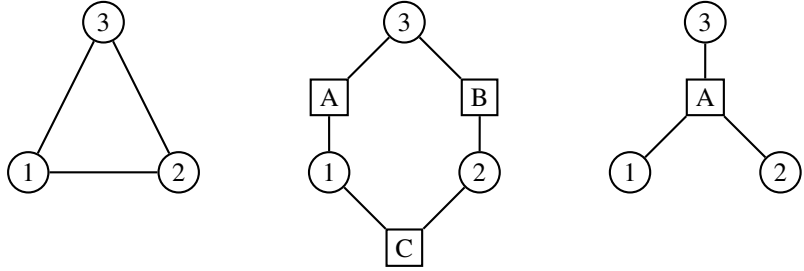
\begin{figure}  
\begin{center}  
\begin{pspicture}(0,0)(11,3)  

\rput(0,1){\circlenode{P1}{1}}  
\rput(2,1){\circlenode{P2}{2}}  
\rput(1,3){\circlenode{P3}{3}}

\ncline{-}{P1}{P2}  \ncline{-}{P2}{P3}  \ncline{-}{P3}{P1}

\rput(4,1){\circlenode{P1}{1}}  
\rput(6,1){\circlenode{P2}{2}}  
\rput(5,3){\circlenode{P3}{3}}
\rput(4,2){\rnode{A}{\psframebox{A}}}
\rput(6,2){\rnode{B}{\psframebox{B}}}
\rput(5,0){\rnode{C}{\psframebox{C}}}

\ncline{-}{P1}{A}  \ncline{-}{P3}{A}
\ncline{-}{P2}{B}  \ncline{-}{P3}{B}
\ncline{-}{P1}{C}  \ncline{-}{P2}{C}  

\rput(8,1){\circlenode{P1}{1}}  
\rput(10,1){\circlenode{P2}{2}}  
\rput(9,3){\circlenode{P3}{3}}
\rput(9,2){\rnode{A}{\psframebox{A}}}

\ncline{-}{P1}{A}  \ncline{-}{P2}{A}  \ncline{-}{P3}{A}

\end{pspicture} 
\caption{Left) Interaction graph of variables corresponding to a problem with variables $X_1$, $X_2$, and $X_3$. Center) Factor graph representing three pair-wise interactions among the variables. Right) Factor graph showing one three-variate interactions.}  
\label{FACT_G}  
\end{center}  
\end{figure}

 One limitation of interaction graphs is that they are not expressive enough to show the order of the interactions among the variables. This is illustrated in Figure~\ref{FACT_G} Left) where we can not discern from the graph whether there are three pair-wise interactions, without a higher three-order contribution, or whether the three variables jointly interact. 

 A \emph{factor graph} \cite{Kschischang_et_al:2001} is a bipartite graph that can serve to represent the factorized structure of a distribution. It has two types of nodes: variable nodes (represented as a circle), and factor nodes (represented  as a square). In the graphs, factor nodes are represented by capital letters starting from $A$, and variable nodes by numbers starting from $1$. Variable nodes are indexed with letters starting with $i$, and factor nodes with letters starting with $a$. The existence of an edge connecting variable node $i$ to factor node $a$ means that $x_i$ is an argument of function $f_a$ in the referred factorization.

 Figure~\ref{FACT_G} Center) and~Figure~\ref{FACT_G} Right) show the way in which factor graphs provide more complete information about the interactions among the variables of a problem. They represent two different patterns of interactions among the variables that would have an identical representation using an interaction graph. 

 Despite the more expressive nature of a factor graph to represent the problem structure, its adoption to model variable interactions in EAs has been limited \cite{Helmi_et_al:2014,Lima_et_al:2009,Mendiburu_et_al:2007a,Muhlenbein:2012}, probably due to the additional number of factor nodes it requires to model a factorization.

\subsection{Modeling the search}

 The schema theory, the most extended formalism used to explain GAs, starts from the assumption that GAs change the sampling rates of hyperplanes in an n-dimensional hypercube corresponding to a binary encoding of the solution space \cite{Goldberg:1989,Holland:1975}. A schema is simply a hyperplane in the search space. It is represented by a chain of symbols taken from the variables domain plus a ``don't care'' symbol (usually $\#$). Two features are used to describe a schema. Its order (number of positions in the schema that do not have the $\#$ symbol) and its defining length (distance between the outermost defined positions). Building blocks are  defined as short, low-order schemata.

The schema theory tries to explain the behavior of algorithms that are defined by the application of crossover and mutation operators. It is not straightforward enough to apply it to explain FDAs, where the graphical representation of the interactions among the variables provides a more flexible encoding of the epistatic nature of the problem. For example, the role of the ``defining length'' of a schema is, to the very least, secondary for the analysis of FDAs, because these algorithms can guarantee that proper  building blocks will not be disrupted during the sampling step,  whatever the position of the variables in the solution representation. As long as the relevant dependencies of the problem are sufficiently covered by the probabilistic model, the behavior of FDAs will not be harmed by the position of the variables in the representation. This is an important difference with GAs, for which the defining length of the schemata can play a primary role to explain the behavior of the crossover operators. 

The defining length is  one of the conceptual components of the schema formalism that become irrelevant for the analysis of FDAs. There are also conceptual gaps, or missing components, in the schema theory that prevent their application to explain FDAs. 

 The limitations of the schema theory to explain EDAs have been analyzed in detail in \cite{Santana_et_al:2007f} and an in-depth analysis of this question is beyond the scope of this paper. The significant point here is whether the key concepts used in schema theory (e.g., hyperplanes, schemata, building blocks, etc.)  are appropriate to explain, and eventually further develop the work on  gray-box optimization. On one hand, gray-box optimization builds on previous work on GAs, proposing ways to enhance traditional recombination operators (e.g., by introducing the partition crossover \cite{Tinos_et_al:2015}). On the other hand, in order to take advantage  of the variable interactions, gray-box optimization relies heavily on the use of a graphical representation, and increasingly on concepts and algorithms defined for them \cite{Whitley_et_al:2016}. The dilemma lies in the fact that some of these graphical-model related concepts and methods have difficulty fitting into the theoretical schema analysis framework traditionally used to analyze GAs.

 By selecting the theory of probabilistic graphical models as a framework to develop EDAs, it was possible to find a more precise  way to explain the behavior of algorithms that exploit problem interactions than recurring to the schema theory. The theory of PGMs also makes it easier to develop these algorithms, since factors and factorizations are taken into account in the design of the algorithms. The position defended in this paper is that work on gray-box optimization illustrates a phenomenon in which competing distinct analytical characterizations of how EAs work collide
 in providing a coherent framework to investigate this type of algorithms.
 
  While a framework that integrates the concepts from the schema theory and those commonly used to explain and design EDAs is not available, some understanding of the relationship between terms deployed in the two areas is needed in order to identify the links between the algorithms. A pair of terms that serve to illustrate this matter is (hyperplanes, factors). 

\subsubsection{Hyperplanes and factors}

 As mentioned before, the analysis of \emph{hyperplanes} has been traditionally used to explain the behavior of GAs and the influence of recombination operators in the search for optimal solutions. Hyperplanes are usually assigned a mean fitness value computed by using all the solutions that belong to the hyperplane. To illustrate the concept, let us use the definitions introduced in  \cite{Whitley:2015} to present the concept of \emph{order-j deceptive} function.

 ``Let  $h$ denote a $(n-j)$-dimensional hyperplane where $j$ variables have preassigned bit values, and $\alpha(h)$ be a mask with $1$ bits marking the locations where the $j$ variables appear in the problem location and $0$ elsewhere.  Let $MAX({\bf{x}},\alpha(h))$ return the hyperplane with the best mean over all $2^j$ order $j$ hyperplanes that can be defined using the  $\alpha(h)$ mask. A function is \emph{order-j deceptive}~\cite{Whitley:2015} if the $j$ bit values returned by  $MAX({\bf{x}},\alpha(h))$ for all hyperplanes of order $j$  are not the same as the bit values found in a string which is a global optimum.''


In the explanations of EDAs, \emph{factors} play a role similar in importance to that played by  hyperplanes in the most common theories used to explain GAs.  But there is not a complete match  between the terms due to the difference between the algorithms. Basically, a factor is a subset of variables from the problem that are modeled together. Usually, the variables are grouped in a factor to indicate that there is some sort of interaction among them in the problem. When a probabilistic model is learned, marginal probabilities are learned for each factor. A marginal probability assigns a probability value to each possible joint configuration of the variables in the factor.  Ideally, in the  marginal probability of the factor, the configurations of the variables with the highest probability will correspond to those with the highest fitness contribution to the global solution. 

In terms of hyperplanes, we can see each of the $2^j$ configurations of a factor $(X_1,\dots,X_j)$ as a hyperplane, in which the marginal value of the configuration serves as an estimate of the corresponding fitness of the hyperplane. Since the table of marginal distributions contains the marginal values for all  $2^j$ configurations, we can easily determine the factor configuration with the best fitness, similarly to the way in which the above-mentioned function $MAX({\bf{x}},\alpha(h))$ returns the hyperplane with the best mean over all $2^j$ order $j$ hyperplanes. Whether the marginal probability is an accurate estimate of the fitness depends on the population size. Assuming an infinite population size, the estimate can be exact. 

 What makes factors exceptionally practical from the algorithm design point of view  is that they can be conveniently combined to form factorizations. As a rule, factorizations are understood as factorizations of a distribution, but factorizations can also be interpreted in more general terms, as possible ways to represent the problem decomposition (see discussion in Section~\ref{sec:PROBDECOMP}). In EDAs, a factorization can serve to represent the way in which the problem should be divided into sets of interacting variables, and equally important, \emph{how these sets should relate to each other} in order to efficiently generate new solutions. Notice that the question of how to effectively encode (reorder) the sets of interacting variables in the hyperplane representation  to increase the efficiency of gray-box optimizers  is related to the issue  of how to  ``factorize'' the representation according to the problem interactions.  One of the tricky issues in the analysis of hyperplanes is the scenario where different hyperplanes overlap. This is elegantly solved in FDAs by means of the graphical representation of the factors, and the methods to sample solutions working on this representation \cite{Muhlenbein_et_al:1999}. 

 Factorizations can be grouped into classes according to the way factors are related, for instance in \cite{Muhlenbein_et_al:1999,Santana:2005,Santana_et_al:2007f} the classes of valid and invalid factorizations are described. In general, decomposable problems have simple factorizations from which it is feasible to compute the best configuration in a short time. However,  more difficult problems can also be approached from the point of view of factorizations.

 While the way in which gray-box optimization and FDAs use the information about the structure of the problem is different, the implications that the characteristics of the graphical representation of the problem structure have for the complexity of the optimization problem are very similar for the two approaches. Decomposable problems are as trivial for FDAs \cite{Muhlenbein_et_al:1999}, as order-k separable Mk landscapes  are for a gray-box optimizer \cite{Whitley_et_al:2016}. A more challenging question in the study of factorizations is the analysis of those problems that are not trivial, their identification, and the specific approaches conceived  to deal with them by exploiting the problem structure. In Section~\ref{sec:GRAPH_DIFFICULTY} we discuss this question. 


\subsubsection{Other representations}

 In addition to hyperplanes and factors, other formalisms have been proposed to model the relationships among the variables in the design of EAs. One of these formalisms is the family of subsets (FOS) \cite{Thierens_and_Bosman:2011}. A FOS is a set of subsets of a certain main set $S$, i.e., a subset of the power-set of $S$. When used to analyze or implement EAs  \cite{Bosman_and_Thierens:2012,Thierens:2011,Thierens_and_Bosman:2011,Thierens_and_Bosman:2013,Tung_and_Yu:2015}, the set $S$ comprises all the problem variables indices. Thus, a FOS actually represents  different subsets of the problem variables, and the choice of these subsets may be related to the problem structure or to the relationships imposed by the algorithm (e.g., a particular way to factorize the problem). 

 Although a subset of indices belonging to a FOS can be interpreted as a factor, and the idea of FOS is similar to the way in which region based approximations are defined in statistical physics to approximate the free energy \cite{Kikuchi:1951,Yedidia_et_al:2002a}, the  FOS representation is very flexible since a certain FOS can be seen independently of the mechanism used to generate it. Algorithms can be characterized by the properties of the FOS they generate. Kikuchi approximations \cite{Hoens:2012}, also used to represent sets of subsets of variables is EDAs, are defined by a specific algorithm that generates them. While FOS have been used to analyze EDAs and hybrid algorithms with characteristics of GAs and EDAs (e.g., linkage-tree GA \cite{Thierens:2011}), as a formalism it is closer to factors and factorizations than to hyperplanes. 



\subsection{Using the structure for the search}

In addition to providing an understanding of the patterns of relationships among the variables, graphical representations are handy for designing new operators and improving the components of EAs. Among the ways the structure can be used to improve the performance of EAs are:

\begin{itemize} 
 \item To bias the learning of the probabilistic models. 
 \item To design more efficient local search and mutation operators.
 \item To design new, hybrid, search approaches. 
 \item To implement transfer learning strategies. 
\end{itemize}

\subsubsection{Using the structure to bias the learning of the probabilistic model}

  Perhaps the most widely employed procedure to exploit the graphical structure of a problem in EDAs is to bias or constrain the way a probabilistic model is learned  \cite{Baluja:2006,Hauschild_et_al:2012,Santana:2005,Schwarz_and_Ocenasek:2000}. This use of the structure is slightly different from the way in which FDAs use the graphical structure. In the latter case, the structure is employed to build a factorization of the problem that will be fixed along the evolution \cite{Santana_et_al:1999}. In the former case, however, a probabilistic model is learned in each generation. The graphical structure of the problem helps to narrow the space of probabilistic models that are searched \cite{Baluja:2006,Muehlenbein_and_Mahnig:2002,Santana:2005}, or to bias the learning process in such a way that those models that capture features of the known problem structure will be more likely to be learned \cite{Hauschild_et_al:2012,Pelikan_et_al:2011,Schwarz_and_Ocenasek:2000}.



\subsubsection{Design of more efficient search operators}

  Designing more efficient search operators is the core idea of local-search gray-box optimization algorithms, such as the hill climber algorithms proposed in \cite{Chicano_et_2014,Chicano_et_al:2016}, which use information about the structure of the function. In the hill climber approach it is assumed that the problem is k-bounded Pseudo-Boolean, and that the number of potential improving moves to be applied by the local optimizer is drastically reduced by taking into account the interactions among the variables. The influence of the interaction among the variables, as captured by the graph structure, is implicitly contemplated in the implementation of the method.  As stated in \cite{Whitley:2015}, if it is assumed that if there are no more single-bit improving bit-flips, then the only pairs of variables that could be mutated together in order to achieve an improvement in the fitness of the function are those linked in the interaction graph.

 
The use of the problem structure is also at the very core of EDAs that use the information about the interactions among the variables to sample them together, instead of independently \cite{Larranhaga_and_Lozano:2002}. Furthermore, a number of proposals of structure-aware mutation operators, and other methods that exploit knowledge about the local structure of the interactions have been proposed in EDAs. Among these proposals are:

\begin{itemize} 
 \item  Structure-aware mutation operators via probabilistic model building of neighborhoods \cite{Sastry_and_Goldberg:2004a}.
 \item Using global statistical information to implement informed mutation operators such as guided mutation \cite{Zhang_and_Sun:2006,Zhang_et_al:2005a}.
 \item  Substructural neighborhoods for local search in the Bayesian optimization algorithm \cite{Lima_et_al:2006,Luong_et_al:2010}.
 \item  Using  belief propagation methods to exchange information about the best local configurations for each set of interacting variables \cite{Lima_et_al:2009,Mendiburu_et_al:2007a,Mendiburu_et_al:2012}.  
 \end{itemize} 

   Efficiency can also be improved by using the structural information as a way to initialize the solutions or populations previous to the application of the genetic operators. In \cite{Hains_et_al:2013},  an algorithm is introduced that uses configurations of variables that correspond to hyperplanes with high quality average fitness to  initialize stochastic local search for MAXSAT.


\subsubsection{Hybrid methods that exploit problem structure}

  Besides gray-box optimization algorithms and EDAs, there are other methods that model the structure of the problem and use the models to generate new solutions. Among these methods are Optimal Mixing EAs (OMEAs) \cite{Thierens_and_Bosman:2011}. We consider these methods as ``hybrid'' because they learn graphical representations using machine learning algorithms similar to those traditionally used by EDAs but, instead of generating new solutions using probabilistic sampling methods, they apply operators reminiscent of those used by a traditional GA.

  A distinguished feature of the different variants of OMEAs is the intermediate evaluation of solutions during the  generation step. Solutions are created by making partial changes to an initial template (e.g., a solution from the population).  Every time that  a solution  is partially modified, it will be evaluated to ascertain whether the modification made over the previous solution was beneficial. At the end of the process, the acceptance of the offspring in the new population will depend on whether any fitness improvement has been produced.  The sets of variables that are modified together at each step of the solution generation are selected according to a ``linkage'' model. The values assigned to this set of values come from another ``donor'' solution that can be selected following different criteria. In the original Linkage Tree GA (LTGA) \cite{Thierens:2011}, mixing is restricted to two parents, in the Genepool Optimal Mixing EA (GOMEA) \cite{Thierens_and_Bosman:2011}, the values selected for each factor of variables come from a random individual. 

 Another characteristic of the OMEAs is the way in which the structural information about the problem is represented. One of the most used models is the  linkage tree, which is a structure commonly used to represent hierarchical clusters. The bottom level of the structure contains $n$ single sets, and in every other level the two most similar sets in the previous level are clustered together.  This multilevel representation adds a different perspective to the way in which structural information is represented. It allows to capture some hierarchical structural relationships among the problem components. However, this convenient property of linkage tree models  is obtained at the expense of not being able to represent problems with overlapping structure. The Linkage Neighbor GOMEA \cite{Bosman_and_Thierens:2012} attempts to address this limitation by modeling, for each variable, the nearest neighbors in terms of linkage (i.e., similar to the concept of  Markov blanket in Markov networks). 

  The large corpus of work on OMEAs  \cite{Bosman_and_Thierens:2012,Bosman_and_Thierens:2013,Goldman_and_Tauritz:2012,Sadowski_et_al:2016,Thierens_and_Bosman:2011,Wang_et_al:2014} provides a set of valuable lessons for the use of the problem structure in EAs, among them we identify the following:

 \begin{itemize}
   \item Parsimonious construction of solutions, with a monotonic improvement of the fitness and strong criterion for acceptance in the new generation, arise as a feasible alternative to probabilistic sampling,  and as an effective way of keeping the population size small. It has been claimed that OM is a superior combination of local search and EAs \cite{Bosman_and_Thierens:2011}.
   \item The uses given to the  selected solutions at the time of model selection can be various. Contrary to the usual practice in EDAs, in LT-GOMEAs, selected solutions are used to learn the model structure \cite{Thierens_and_Bosman:2013}, but they are not used in any way for biasing  the assignments of the variables values.  
   \item  The efficiency of the OMEAs depends on the choice of the FOS or linkage sets, and on the relationship between these linkage sets and the problem structure \cite{Wang_et_al:2014}.  LTGA does not seems to perform well for 2D spin glasses \cite{Pelikan_et_al:2011} and problems with complex overlapping linkage structure \cite{Goldman_and_Tauritz:2012}.
   \item In terms of the implications that knowledge about the problem structure may have for the search, perhaps the main message from the application of GOMEAs is as stated in \cite{Thierens_and_Bosman:2011}: ``This supports the notion that it is not just finding a good configuration of a sufficiently complex structure, it is also the way in which this structural information is exploited upon creating new solutions that is of vital importance.''
 \end{itemize}


\subsubsection{Transfer learning strategies based on structural similarity}

 A natural extension of the use of the graphical structure of a given problem is to exploit this information for problems that are different but share some structural characteristics. This is the case, for example, of different instances of the same problem (e.g., Max-SAT, NK-landscape, etc.). In these scenarios  it would be desirable that EAs exploit information about the structure of one (source) problem at the time of optimizing other (target) problems.  In \cite{Pelikan_and_Hauschild:2012} a transfer learning approach that computes a  distance-based metric to bias the learning of the probabilistic model while solving a target problem instance is proposed. The metric is computed  from the Bayesian network structures learned by hBOA while solving other source instances of the same problem. In  \cite{Santana_et_al:2012f}, different variants of structural transfer are discussed. For the application examined it is shown that the use of partial information extracted from a set of source problems reduces the computational cost associated with probabilistic model learning and improves the quality of the final results for the target problem.

\subsection{Problem structure, problem difficulty, and algorithm behavior}  \label{sec:GRAPH_DIFFICULTY}

 One initial consideration that we make is the difference between the general question of what makes  a problem difficult for EAs, as investigated in \cite{Forrest_and_Mitchell:1993,Manderick_et_al:1991,Mitchell_et_al:1992a,Thierens_et_al:1998}, and the more specific question of how the structure of a problem influences its  difficulty for an EA. The first question is very broad since several factors can make a problem difficult for EAs. 

  Although more specific, the question of how the problem structure  influences the difficulty that the problem poses to an EA is itself very complex and beyond the scope of this paper. Here we only sketch some points related to the problem structure and their impact on the results of FDAs and gray-box optimization algorithms. Our main claim here is that there are structural features that have a very similar impact on the two approaches. Indeed, most of these features can be derived from the analysis of the problem graphical structure. 

\subsubsection{Trivial and hard functions for gray-optimization and FDAs}

 k-order separable decomposable functions are easy both for gray-box optimization methods \cite{Whitley_and_Chen:2012} and FDAs  \cite{Muhlenbein_et_al:1999}. The complexity for both algorithms is exponential in $k$.  A number of recent results in gray-box optimization are related to this fact \cite{Whitley:2015,Whitley_et_al:2016}.  In \cite{Whitley:2015},  a  pre-processing algorithm has been introduced that solves all separable Mk landscapes problems in $O(n)$ time. It has been also shown that ``Localized Mk landscapes'' can be constructed that are not adjacent, but which also can be solved in polynomial time.


 Two findings shape the discussion on the relationship between the structural or graphical characteristics of a problem and its difficulty for FDAs. The first result,  \cite{Muhlenbein_et_al:1999}, points to the identification of the tree-width of a junction tree (the size of the largest node in the  junction tree minus 1) constructed from the ADF structure as an indicator of problem complexity for FDA. This result was a breakthrough since it allowed to establish a connection between GA analysis, the theory of probabilistic graphical models, and well-known dynamic programming algorithms.

 A second, very influential, result was presented by Gao and Culberson in \cite{Gao_and_Culberson:2005}. They showed that for random ADFs the tree-width of the corresponding graphical structure is of order $n$.  While k-order separable ADFs have tree-width $(k-1)$ and can be solved in $O(n)$,  random ADFs of order $k$ are exponentially complex for FDAs. Interestingly, EDAs that learn Bayesian networks can also see their performance quickly deteriorate for problems with an increasing number of $k$-order subfunctions \cite{Echegoyen_et_al:2011}. Recently, the connection between the work  Gao and Culberson \cite{Gao_and_Culberson:2005} and the complexity of ADFs for gray-box optimization problems has been discussed in \cite{Whitley_et_al:2016}. As expected, random functions are also exponentially complex for gray-box optimization algorithms.

 While results for random functions point to the limits of optimization methods that use the problem structure, real-world optimization problems are not random in general. Therefore, the essential consideration at the time of addressing one real-world problem is what is the tree-width of its associated structure graphical representation.  Nonetheless, even if some features that characterize the problem difficulty for FDAs and gray-box optimization methods can be extracted from the analysis of its graphical structure, the extraction of these features is not always straightforward. For example, the problem of finding the triangulation of a graph with minimum tree-width problems is NP-hard \cite{Yannakakis:1981}. A positive note is that there are fast algorithms which are capable of computing approximations \cite{Heggernes:2006,Tarjan_and_Yannakakis:1984}.

 Another question that is usually overlooked in the analysis of the problem structure is the type and strength of the interactions among the variables.  In \cite{Santana_et_al:2005b}, capturing benign and malign interactions \cite{Kallel_et_al:2000} in the model is shown to have a different effect on the behavior of EDAs.  Similarly, weak and strong interactions are not equally significant at the time of modeling a problem. Therefore, while an analysis based on the tree-width of the graphical representation  and on other structural characteristics provides a global picture of the complexity of the problem, a detailed picture will depend on the type and strength of the interactions represented by the structure.



%


 \section{Competing  analytical characterizations at work}   \label{sec:GRAPHS}
 
  In this section we present an example that illustrates how an analytical approach based on the schema theory can be misleading at the time of using and understanding how the problem structure influences the behavior of the algorithms. 

\subsection{A link between hyperplane statistics and marginal distributions}

\subsubsection{Boltzmann distribution}

 The Boltzmann probability distribution  $p_B({\bf{x}})$ (also called Gibbs distribution) is defined as:  

  \begin{equation}
  p_B({\bf{x}}) = \frac{e^{\frac{g({\bf{x}})}{T}}}{
  \sum_{{\bf{x}}'} e^{\frac{g({\bf{x}}')}{T}}}, \label{eq:BOLTPROB}
  \end{equation}
  where $g({\bf{x}})$ is a given objective function and $T$ is the system temperature that can be used as a parameter to smooth the probabilities. The inverse temperature  $\beta = \frac{1}{T}$ is frequently used instead of the temperature. 

A convenient property of $p_B({\bf{x}})$ is that it assigns a higher probability to solutions with better fitness. The solutions with the highest probability correspond to the optima.  The relationship between the fitness function and the variables dependencies that arise in the selected solutions can be modeled using the Boltzmann probability distribution  \cite{Hoens:2005,Mahnig:2001b,Muehlenbein_and_Mahnig:2002a,Muhlenbein_et_al:1999}. 

\subsection{The case of problems with circular structure}

 To illustrate how an analysis based on hyperplanes can be misleading, we take an adjacent MK landscape presented in \cite{Whitley:2015,Whitley_et_al:2016}. For this function, we replicate the computation of a number of hyperplanes, as done in  \cite{Whitley:2015,Whitley_et_al:2016}, and also compute other factorizations of the problem derived from the analysis of its graphical representation. We then compare the findings from these two approaches.

 The additive function has  $n=10$ variables and is computed as the sum of $10$ subfunctions of order $k=3$.  Table~\ref{tab:FUNCTION_WHITLEY} presents the definition as taken from \cite{Whitley:2015}. Each subfunction has four local optima with evaluation $1$ and all other points have evaluation $0$. For each subfunction, one of the four local optima is reached at $111$. The global optimum of the function is reached at a vector with all variables set to $1$. 

\begin{table}[htb]
\scriptsize
\begin{center}
{$\begin{array}{c|c|c c c c|c} \hline \hline
  \text{Index}  & \text{subfunction} & \multicolumn{4}{c|}{\text{local optima}}  & \text{codomain vector}  \\  \hline  \hline
  1 &  f(x_1,x_2,x_3)  &   000 & 001 & 100 & 111 & <1,1,0,0,1,0,0,1> \\ 
  2 &  f(x_2,x_3,x_4)  &   011 & 001 & 100 & 111 & <0,1,0,1,1,0,0,1> \\ 
  3 &  f(x_3,x_4,x_5)  &   001 & 010 & 101 & 111 & <0,1,1,0,0,1,0,1> \\ 
  4 &  f(x_4,x_5,x_6)  &   110 & 100 & 011 & 111 & <0,0,0,1,1,0,1,1> \\ 
  5 &  f(x_5,x_6,x_7)  &   001 & 100 & 110 & 111 & <0,1,0,0,1,0,1,1> \\ 
  6 &  f(x_6,x_7,x_8)  &   000 & 100 & 010 & 111 & <1,0,1,0,1,0,0,1> \\ 
  7 &  f(x_7,x_8,x_9)  &   001 & 010 & 101 & 111 & <0,1,1,0,0,1,0,1> \\ 
  8 &  f(x_8,x_9,x_0)  &   000 & 001 & 100 & 111 & <1,1,0,0,1,0,0,1> \\ 
  9 &  f(x_9,x_0,x_1)  &   000 & 010 & 011 & 111 & <1,0,1,1,0,0,0,1> \\ 
  10&   f(x_0,x_1,x_2)  &   000 & 101 & 110 & 111 & <1,0,0,0,0,1,1,1> \\ \hline  \hline
    \end{array}$}
\caption{Example of adjacent MK landscape presented in \cite{Whitley:2015}. Each subfunction has four local optima with evaluation $1$; all other points have evaluation $0$. These values are represented by the codomain vector. The global optimum is the string of all $1$ bits, and each subfunction reaches the optimum at $111$.}
\label{tab:FUNCTION_WHITLEY}
\end{center}
\end{table}

 We then generate the complete space of $2^{10}=1024$ solutions and evaluate them. Using the solutions and their corresponding evaluations, it is possible to compute the average fitness of any partial configuration of any subset of variables (factor). We start by computing the marginal frequencies for factors corresponding to all definition sets of the function, i.e., those triplets of variables shown in column $2$ of Table~\ref{tab:FACTORS_J}.

\begin{table}[htb]
\scriptsize
\begin{center}
{$\begin{array}{r|c|c|c} \hline \hline
  \text{Index}  & j=3 & j=4 & j=5  \\  \hline  \hline
  1 &  (x_1,x_2,x_3)  &  (x_1,x_2,x_3,x_4)  &  (x_1,x_2,x_3,x_4,x_5)    \\ 
  2 &  (x_2,x_3,x_4)  &  (x_2,x_3,x_4,x_5)  &  (x_2,x_3,x_4,x_5,x_6)      \\ 
  3 &  (x_3,x_4,x_5)  &  (x_3,x_4,x_5,x_6)  &  (x_3,x_4,x_5,x_6,x_7)  \\ 
  4 &  (x_4,x_5,x_6)  &  (x_4,x_5,x_6,x_7)  &  (x_4,x_5,x_6,x_7,x_8)    \\ 
  5 &  (x_5,x_6,x_7)  & (x_5,x_6,x_7,x_8)  & (x_5,x_6,x_7,x_8,x_9)    \\ 
  6 &  (x_6,x_7,x_8)  &  (x_6,x_7,x_8,x_9)  &  (x_6,x_7,x_8,x_9,x_{0})     \\ 
  7 &  (x_7,x_8,x_9)  &   (x_7,x_8,x_9,x_0)  &  (x_7,x_8,x_9,x_0,x_1)    \\ 
  8 &  (x_8,x_9,x_0)  &  (x_8,x_9,x_0,x_1)  & (x_8,x_9,x_0,x_1,x_2)  \\ 
  9 &  (x_9,x_0,x_1)  &  (x_9,x_0,x_1,x_2)  &  (x_9,x_0,x_1,x_2,x_3)    \\ 
  10&   (x_0,x_1,x_2)  &  (x_0,x_1,x_2,x_3)  &    (x_0,x_1,x_2,x_3,x_4)  \\ \hline  \hline
    \end{array}$}
\caption{Different sets of factors used to compute the marginal frequencies of the function defined in Table~\ref{tab:FUNCTION_WHITLEY}.}
\label{tab:FACTORS_J}
\end{center}
\end{table}

 Since all factors in a given column have the same size, instead of calculating the average fitness, we simply add the fitness values of all solutions where a specific configuration of the factor is present. For example, for column $2$, where $k=3$ and the number of solutions of the search space that share  a given configuration $(x_i,x_{i+1},x_{i+2})$ is $2^{n-k}=2^7=128$, we do not divide the sum of the frequencies for $(x_i,x_{i+1},x_{i+2})$ by $128$.  The rank of the frequencies for each factor will coincide with the rank of the statistics computed for the corresponding hyperplanes of order $k$.  For $k=3$,  the results are shown in Table~\ref{tab:NKPF}, where, for each factor, the configurations with the highest marginal frequency are underlined.  
 
\begin{table}[htb]
\scriptsize
\begin{center}
{$\begin{array}{c||r|r|r|r|r|r|r|r|r|r} \hline \hline
  &  \multicolumn{10}{|c}{\text{Frequencies for the subfunctions}} \\  \hline  
     &  1   &  2   &  3   & 4    &  5   &  6   & 7    &  8   & 9    & 10 \\  \hline  \hline
000  & \underline{768}  & \underline{704}  & 608  & 480  & 480  & 544  & \underline{704}  & \underline{736}  & 640  & 736 \\
001  & 512              & \underline{704}  & \underline{800}  & 736  & 416  & 672  & 576  & 672  & \underline{768}  & 544 \\
010  & 512                          & 576  & 544  & 672  & 608  & 480  & \underline{704}  & 672  & 576  & 672 \\ 
011  & 512                          & 576  & 736  & 672  & \underline{800}  & 480  & 576  & 480  & 704  & 736 \\
100  & 640              & \underline{704}  & 608  & 416  & 736  & 736  & \underline{704}  & 672  & 640  & 544 \\
101  & 640                          & 576  & 544  & 672  & 544  & 608  & 576  & 608  & 640  & 480 \\
110  & \underline{768}              & 576  & 544  & 608  & 736  & \underline{800}  & 576  & 608  & 448  & 608 \\
111  & \underline{768}  & \underline{704}  & 736  & \underline{864}  & \underline{800}  & \underline{800}  & \underline{704}  & 672  & 704  & \underline{800} \\ \hline \hline  
    \end{array}$}
\caption{Marginal frequencies for all the configurations of factors of order $3$. These factors correspond to the definition sets of the function presented in Table~\ref{tab:FUNCTION_WHITLEY}. Marginal frequencies are computed as the sum of the fitness for all solutions that contain the corresponding configuration of the factor. In the table, the configurations where the marginal configuration reached the highest value are underlined.}
\label{tab:NKPF}
\end{center}
\end{table}

It can be seen in Table~\ref{tab:NKPF} that among the factor configurations  with highest value, there is always one that agrees with the global optima of the function. This happens for all factors except in three of them: $3$, $8$, and $9$. In some way these hyperplanes are deceptive. In the vein of the analysis presented in  \cite{Whitley:2015}, we expand the hyperplanes to include more adjacent variables as shown in columns $3$ and $4$ of Table~\ref{tab:FACTORS_J}. The results of the computation of the marginal frequencies are displayed in Table~\ref{tab:NKPF1} and Table~\ref{tab:NKPF2}, respectively.

\begin{table}[htb]
\begin{center}
\scriptsize
{$\begin{array}{c||r|r|r|r|r|r|r|r|r|r} \hline \hline
     &  \multicolumn{10}{|c}{\text{Frequencies for the subfunctions}} \\  \hline  
     &  1   &  2   &  3   & 4    &  5   &  6   & 7    &  8   & 9    & 10 \\  \hline  \hline
0000  & 384&   304&   240&   256&   208&   304&   \underline{368}&   336&   368&   \underline{432} \\
0001  & 384&   \underline{400}&   368&   224&   272&   240&   336&   \underline{400}&   272&   304 \\
0010  & 256&   304&   400&   320&   208&   368&   336&   304&   368&   272 \\
0011  & 256&   \underline{400}&   400&   416&   208&   304&   240&   368&   \underline{400}&   272 \\
0100  & 288&   304&   208&   384&   336&   272&   \underline{368}&   336&   304&   336 \\
0101  & 224&   272&   336&   288&   272&   208&   336&   336&   272&   336 \\
0110  & 224&   240&   304&   320&   \underline{400}&   208&   272&   176&   304&   368 \\ 
0111  & 288&   336&   \underline{432}&   352&   \underline{400}&   272&   304&   304&   \underline{400}&   368 \\ 
1000  & 320&   304&   240&   224&   336&   400&   \underline{368}&   304&   368&   336 \\ 
1001  & 320&   \underline{400}&   368&   192&   \underline{400}&   336&   336&   368&   272&   208 \\  
1010  & 320&   240&   272&   288&   272&   336&   336&   272&   304&   240 \\ 
1011  & 320&   336&   272&   384&   272&   272&   240&   336&   336&   240 \\
1100  & \underline{416}&   304&   208&   352&   \underline{400}&   \underline{432}&   304&   304&   240&   304 \\
1101  & 352&   272&   336&   256&   336&   368&   272&   304&   208&   304 \\
1110  & 352&   304&   304&   416&   \underline{400}&   368&   336&   272&   304&   400 \\
1111  & \underline{416}&   \underline{400}&   \underline{432}&   \underline{448}&   \underline{400}&   \underline{432}&   \underline{368}&   \underline{400}&   \underline{400}&   400 \\\hline \hline 
    \end{array}$}
\caption{Marginal frequencies for all the configurations of factors of order $4$ corresponding to the definition sets of the function presented in Table~\ref{tab:FUNCTION_WHITLEY}.}
\label{tab:NKPF1}
\end{center}
\end{table}

 For factors of order $4$ (Table~\ref{tab:NKPF1}), among the factor configurations with the highest value, there is always one that agrees with the global optima of the function for all factors except one: factor $10$. The same happens for factors of order $5$ (Table~\ref{tab:NKPF2}), but in this case the ``deceptive'' factor is $9$.   We point out that the configurations with the highest marginal values for $j=5$  do not agree with those published in  \cite{Whitley:2015,Whitley_et_al:2016} where only for $4$ of the $10$ hyperplanes the configuration $11111$ was among those with the highest value of the hyperplane statistics\footnote{As supplementary information, we have included the code we have used to compute the marginal frequencies for all factors, as well as the output of the algorithms.  This code is available from \url{https://github.com/rsantana-isg/graybox_fda_paper}}. This lack of consistency between the configuration of the global optima and the configuration of the hyperplanes led the authors to consider the function as deceptive at order $5$. The exact quote in \cite{Whitley_et_al:2016} is: ``Despite the fact that this is a simple Adjacent NK Landscape (with localized variable interactions and localized nonlinearity) the order $5$ hyperplane statistics do not provide any recognizable mechanism for identifying the global optimum. This simple function is still deceptive at order $5$.''

\begin{table}[htb]
\begin{center}
\scriptsize
{$\begin{array}{c||r|r|r|r|r|r|r|r|r|r} \hline \hline
     &  \multicolumn{10}{|c}{\text{Frequencies for the subfunctions}} \\  \hline  
     &  1   &  2   &  3   & 4    &  5   &  6   & 7    &  8   & 9    & 10 \\  \hline  \hline
00000& 168 & 120 & 128 & 112 & 120 & 160 & 168 & 192 & \underline{216} & \underline{216}\\
00001& 216 & 184 & 112 & 144 & 88 & 144 & 200 & 144 & 152 &  \underline{216}\\
00010& 168 & 200 & 160 & 112 & 152 & 144 & 152 & 192 & 136 & 152\\
00011& 216 & 200 & 208 & 112 & 120 & 96 & 184 & 208 & 136 & 152\\
00100& 136 & 120 & \underline{224} & 176 & 120 & 192 & 168 & 160 & 184 & 152\\
00101& 120 & 184 & 176 & 144 & 88 & 176 & 168 & 144 & 184 & 120\\
00110& 104 & 168 & 192 & 208 & 88 & 144 & 88 & 160 & 200 & 120\\
00111& 152 & \underline{232} & 208 & 208 & 120 & 160 & 152 & 208 & 200 & 152\\
01000& 120 & 120 & 112 & 176 & 184 & 144 & 168 & 192 & 184 & 168\\
01001& 168 & 184 & 96 & 208 & 152 & 128 & 200 & 144 & 120 & 168\\
01010& 88 & 136 & 144 & 144 & 152 & 128 & 152 & 160 & 136 & 168\\
01011& 136 & 136 & 192 & 144 & 120 & 80 & 184 & 176 & 136 & 168\\
01100& 120 & 88 & 176 & 176 & \underline{216} & 112 & 136 & 96 & 152 & 200\\
01101& 104 & 152 & 128 & 144 & 184 & 96 & 136 & 80 & 152 & 168\\
01110& 120 & 136 & 208 & 176 & 184 & 128 & 120 & 128 & 200 & 168\\
01111& 168 & 200 & \underline{224} & 176 & \underline{216} & 144 & 184 & 176 & 200 & 200\\
10000 & 136 & 120 & 128 & 96 & 184 & 208 & 168 & 176 & \underline{216} & 168\\
10001 & 184 & 184 & 112 & 128 & 152 & 192 & 200 & 128 & 152 & 168\\
10010 & 136 & 200 & 160 & 96 & \underline{216} & 192 & 152 & 176 & 136 & 104\\
10011 & 184 & 200 & 208 & 96 & 184 & 144 & 184 & 192 & 136 & 104\\
10100 & 168 & 88 & 160 & 160 & 152 & 176 & 168 & 144 & 152 & 136\\
10101 & 152 & 152 & 112 & 128 & 120 & 160 & 168 & 128 & 152 & 104\\
10110 & 136 & 136 & 128 & 192 & 120 & 128 & 88 & 144 & 168 & 104\\
10111 & 184 & 200 & 144 & 192 & 152 & 144 & 152 & 192 & 168 & 136\\
11000 & 184 & 120 & 112 & 160 & \underline{216} & \underline{224} & 136 & 176 & 152 & 152\\
11001 & \underline{232} & 184 & 96 & 192 & 184 & 208 & 168 & 128 & 88 & 152\\
11010 & 152 & 136 & 144 & 128 & 184 & 208 & 120 & 144 & 104 & 152\\
11011& 200 & 136 & 192 & 128 & 152 & 160 & 152 & 160 & 104 & 152\\
11100& 184 & 120 & 176 & \underline{224} & 216 & 192 & 168 & 144 & 152 & \underline{216}\\
11101& 168 & 184 & 128 & 192 & 184 & 176 & 168 & 128 & 152 & 184\\
11110& 184 & 168 & 208 & \underline{224} & 184 & 208 & 152 & 176 & 200 & 184\\
11111& \underline{232} & \underline{232} & \underline{224} & \underline{224} & \underline{216} & \underline{224} & \underline{216} & \underline{224} & 200 & \underline{216}\\ \hline \hline 
    \end{array}$}
\caption{Marginal frequencies for all the configurations of factors of order $5$ corresponding to the definition sets of the function presented in Table~\ref{tab:FUNCTION_WHITLEY}.}
\label{tab:NKPF2}
\end{center}
\end{table}

As previously mentioned, our computation of the marginal frequencies of the factors does not agree with the results published  in  \cite{Whitley:2015,Whitley_et_al:2016}.  However, in both cases,  in the results presented in \cite{Whitley:2015,Whitley_et_al:2016} as well as in Table~\ref{tab:NKPF2}, not all factors have a maximal configuration that agrees with the global optima. Therefore, from the hyperplane analysis we could label  the function ``deceptive''.

 This example shows that expanding the hyperplanes by considering contiguous variables to detect the degree of deception in functions with overlapping definition sets can be misleading. The mishap is at the time of identifying which are the relevant hyperplanes or factors. The fact that not every factor among the $10$ of order $5$ shown in Table~\ref{tab:FACTORS_J} are consistent with the global optima does not mean that there could be other  factors, of the same order, the combination of which could lead an EA to the optimum. Actually, as we will show in what follows, six factors of order five are sufficient to get an exact factorization for this problem. Therefore, for these factors, the problem is not deceptive.

 The problem of identifying the ``relevant'' hyperplanes can be re-framed as the problem of finding a minimal valid factorization that captures every single interaction of the original function, and this can be achieved by the triangulization of the interaction graph and the computation of a junction tree from the triangulized graph.  Computing this type of factorizations from the problem structure and using them to solve deceptive functions were two of the most important results introduced in \cite{Muhlenbein_et_al:1999} and extensively exploited in EDAs since then. 

 Figure~\ref{fig:INTGRAPH} Right) shows a set of edges that, when added to the original interaction graph Figure~\ref{fig:INTGRAPH} Left), produce a chordal or triangulated graph. Table~\ref{tab:NKPF_FACT} shows the six factors obtained from the triangulated graph as well as the marginal frequencies for all the factor configurations. It can be seen that, for all factors the configuration $11111$, consistent with the global optima, is among those with the highest fitness value. Equation~\eqref{eq:FACTORIZATION} shows the factorization that can be obtained from these six factors. 
  

\begin{table}[htb]
\begin{center}
\scriptsize
{$\begin{array}{c||r|r|r|r|r|r} \hline \hline
     &  \multicolumn{6}{|c}{\text{Factor frequencies}} \\  \hline  
     & (0,1,2,8,9) & (1,2,3,8,9) & (2,3,4,8,9)  & (3,4,5,8,9)   & (4,5,6,8,9)  & (5,6,7,8,9) \\  \hline  \hline
   00000  &  \underline{216}  &  160  &  120  &  120  &  144  &  168  \\  
  00001  &  168  &  160  &  136  &  136  &  160  &  200  \\  
  00010  &  168  &  128  &  88  &  88  &  96  &  152  \\  
  00011  &  152  &  160  &  136  &  136  &  144  &  184  \\  
  00100  &  \underline{216}  &  \underline{208}  &  184  &  104  &  160  &  168  \\  
  00101  &  168  &  \underline{208}  &  200  &  120  &  176  &  168  \\  
  00110  &  168  &  176  &  152  &  72  &  144  &  88  \\  
  00111  &  152  &  \underline{208}  &  200  &  120  &  192  &  152  \\  
  01000  &  152  &  144  &  168  &  152  &  128  &  168  \\  
  01001  &  168  &  144  &  184  &  168  &  144  &  200  \\  
  01010  &  104  &  112  &  136  &  120  &  80  &  152  \\  
  01011  &  152  &  144  &  184  &  168  &  128  &  184  \\  
  01100  &  152  &  192  &  168  &  200  &  112  &  136  \\  
  01101  &  168  &  192  &  184  &  \underline{216}  &  128  &  136  \\  
  01110  &  104  &  160  &  136  &  168  &  96  &  120  \\  
  01111  &  152  &  192  &  184  &  \underline{216}  &  144  &  184  \\  
  10000  &  152  &  144  &  104  &  184  &  192  &  168  \\  
  10001  &  200  &  176  &  120  &  200  &  208  &  200  \\  
  10010  &  136  &  112  &  72  &  152  &  144  &  152  \\  
  10011  &  \underline{216}  &  176  &  120  &  200  &  192  &  184  \\  
  10100  &  120  &  128  &  168  &  136  &  144  &  168  \\  
  10101  &  168  &  160  &  184  &  152  &  160  &  168  \\  
  10110  &  104  &  96  &  136  &  104  &  128  &  88  \\  
  10111  &  184  &  160  &  184  &  152  &  176  &  152  \\  
  11000  &  120  &  128  &  152  &  184  &  208  &  136  \\  
  11001  &  168  &  160  &  168  &  200  &  \underline{224}  &  168  \\  
  11010  &  104  &  96  &  120  &  152  &  160  &  120  \\  
  11011  &  184  &  160  &  168  &  200  &  208  &  152  \\  
  11100  &  152  &  176  &  216  &  200  &  192  &  168  \\  
  11101  &  200  &  \underline{208}  &  \underline{232}  &  \underline{216}  &  208  &  168  \\  
  11110  &  136  &  144  &  184  &  168  &  176  &  152  \\  
  11111  &  \underline{216}  &  \underline{208}  &  \underline{232}  &  \underline{216}  &  \underline{224}  &  \underline{216}  \\ \hline  \hline 
    \end{array}$}
\caption{Marginal frequencies for factorization represented by Equation~\eqref{eq:FACTORIZATION}.}
\label{tab:NKPF_FACT}
\end{center}
\end{table}

\begin{align}
  p({\bf{x}}) &= p(x_0x_1x_2x_8x_9) \cdot p(x_3|x_1x_2x_8x_9)\cdot p(x_4|x_2x_3x_8x_9) \nonumber \\
              &\cdot p(x_5|x_3x_4x_8x_9) \cdot p(x_6|x_4x_5x_8x_9) \cdot  p(x_7|x_5x_6x_8x_9) \label{eq:FACTORIZATION}
\end{align}

In \cite{Whitley_et_al:2016}, it is also proved that functions with circular structure of order $K$, where $K$ is the number of neighbors (cyclic adjacent NK landscapes), can be converted to acyclic adjacent NK landscape of order $2K$.  The function shown in Table~\ref{tab:FUNCTION_WHITLEY} is a cyclic adjacent NK landscape and for a function with the same structure it is also shown in  \cite{Whitley_et_al:2016} how to construct a possible acyclic adjacent NK landscape with $K=4$, i.e., factors of order five.


\section{Improving and extending the scope of methods that exploit the problem structure}  \label{sec:FUTURE}
 To this point we have provided evidence that FDAs and gray-box optimization algorithm benefit from the graphical structure of the problems and that they are similarly constrained in their performance by the properties of these structures. Therefore, in this section we discuss a number of issues that are challenging for these two classes of algorithms, and for other EAs that exploit the problem structure. We also discuss issues that represent an opportunity to enhance and extend these algorithms. The following topics are covered:

 \begin{itemize} 
   \item Problems with large definition sets and a high cardinality of the variables.    
   \item Estimating or learning the structure of black-box problems.   
   \item Redefining problem structure for constrained and multi-objective problems. 
   \item The question of evolvability.
   \item Combining gray-box optimization and EDAs.
 \end{itemize}

\subsection{Large sets of interacting variables and high cardinality of the variables}
  
  One common challenging problem for EDAs and gray-box optimization are problems for which there are large interacting subsets of variables. Usually, FDAs and gray-box optimizers explicitly set constraints on the structural characteristics of the problem. For example, the gray-box optimization method introduced in \cite{Chicano_et_2014} assumes that the number of subfunctions in which any variable appears must be bounded by some constant $c$. Similarly, some EDAs set a limit to the size of the biggest clique in the learned factorization \cite{Santana_et_al:2008c,Santana_et_al:2009c}, or to the maximum number of neighbors a variable may have in the interaction graph  \cite{Shakya_et_al:2011}. The downside of this type of restrictions is that either the algorithms can not be applied to problems that do not respect the structural assumptions, or the behavior of the algorithms  in these cases cannot be predicted. It is worth noticing that in some real-word problems some variables can play a critical role by appearing in a large number of subfunctions \cite{Whitley_et_al:2013}.

 Another problematic issue is the cardinality of the variables. Most gray-box optimization methods and FDAs have been exclusively tested using binary problems. This restriction is doubly limiting, on one hand because this is the lowest cardinality for problems with discrete representation, on the other hand because binary problems also have the cardinality for all variables fixed. Therefore, problems with non-uniform cardinality of the variables are not usually addressed and it is not possible to assess how a non-uniform distribution of the variables cardinalities can impact the behavior of these algorithms. The most difficult scenario is comprised by problems that exhibit the two characteristics, they have large sets of interacting variables and high cardinality of the variables. For these problems it has been shown that even FDAs that use exact factorizations suffer an important impact in their performance compared with the scenario in which binary problems are solved \cite{Santana_et_al:2009c}.

\subsection{Estimating or learning the structure of black-box problems}

 When no information about the problem structure exists, methods that learn  a model from data are usually employed in EDAs. Probabilistic graphical models actually learn a representation of the dependence relationships among the variables and probabilistic patterns (e.g., marginal distributions) in the data. However, it is usually assumed in EDAs that the dependence relationships reflect the problem structure. The link between problem structure and probabilistic dependencies is not clearly defined because it depends on aspects such as the choice of the probabilistic model, the problem representation, the strength of selection, etc.  To complicate the matter more, models that accurately capture the interactions between the problem variables are not always more effective than models that discard or ignore these interactions. Therefore, an algorithm can excel at recovering the problem structure and producing mediocre results in terms of the best solutions found.  

 The study of the link between problem structure and model dependencies is beyond the scope of this paper. The interested reader can consult \cite{Brownlee_et_al:2015,Brownlee_et_al:2012a,Brownlee_et_al:2009,Echegoyen_et_al:2011,Mishra_and_Gallagher:2012,Morgan_and_Gallagher:2012} for different perspectives on this topic. Similarly, an account of the different types of probabilistic models used in EDAs and how they represent the relationships between variables can be obtained from several surveys on EDAs \cite{Hauschild_and_Pelikan:2011,Larranaga_et_al:2012,Lozano_et_al:2005,Shakya_and_Santana:2012a}.  We focus our analysis on the recent renewed interested in the application of neural models in EDAs. 

\subsubsection{Neural and deep learning models in estimation of distribution algorithms}

 The use of neural networks (NNs) as models in EDAs is not new \cite{Marti_et_al:2008,Shim_and_Tan:2012,Tang_et_al:2010,Zhang_and_Shin:2000}. However, an increasing  number of recent works \cite{Baluja:2017,Churchill_et_al:2016,Probst_and_Rothlauf:2015,Probst_et_al:2017,Saikia_et_al:2016}  propose the application of neural  models which, in some cases, are also deep learning architectures.  The particular characteristics of the neural models and the possibility of applying ``deep neural models'' in EDAs motivates the brief review included in this section. The main questions we address are: 1) What is essentially different in the information about the problem structure conveyed by neural models? 2) What advantages and disadvantages exhibit the neural model representations, and NN learning and sampling algorithms, within the context of model-based EAs?

  A fundamental difference of NNs over graphical models is that information about the problem structure is usually represented in latent variables or distributed structures that make interpretability of the model a difficult task. From the perspective of the main topic investigated in this paper, the exploitation of the problem structural information for optimization, the question arises as to what extent a latent representation of the optimization problem can be efficiently exploited. 

 In  classification problems, latent features produced by NN learning methods can play a very important role \cite{Ciresan_et_al:2011}.  In addition to the direct use of the latent features produced by NNs for generating new solutions in the contexts EAs,  there are potential benefits in extracting knowledge about the search space from these latent representations.  An increasing number of methods are being proposed for interpreting neural models \cite{Montavon_et_al:2017}. In particular, methods for the identification of feature interactions of arbitrary order from the analysis of the neural network weights have been recently proposed \cite{Tsang_et_al:2017}. A number of papers that have applied neural models in EDAs have inspected and analyzed the models learned as a source of knowledge about the problem characteristics. Table~\ref{tab:NN_MODELS} summarizes some relevant characteristics of a number of works that introduce neural models.

Although the main evident difference among the approaches shown in Table~\ref{tab:NN_MODELS} is the type of neural network used by the algorithms, significant differences can exist in the particular way the neural models are applied. In traditional EDAs based on  probabilistic graphical models, different methods can be used for learning the model. Perhaps the best known example is the variety of learning algorithms employed by EDAs based on Bayesian networks \cite{Etxeberria_and_Larranhaga:1999,Muehlenbein_and_Mahnig:2001a,Pelikan_et_al:1999}).  Something similar applies for sampling methods in EDAs, where even for a specific Markov network based EDA, different sampling procedures produce significantly distinct  behaviors of the algorithm \cite{Santana_et_al:2013b}. 

In EDAs that use neural models, these differences can be more noticeable because there can be multiple ways to exploit the information contained in a NN. On top of that, the NN can be applied in multiple ways. For instance, autoencoders \cite{Bengio:2009,Vincent_et_al:2010}, a particular class of NNs, have been used to sample new solutions in a way that resembles sampling in EDAs  \cite{Probst:2015a}, but also to adaptively learn a ``non-linear mutation operator'' that attempts to  widen the basins of attraction around the best individuals. In terms of the EDA performance,  the potential of the neural model to capture intricate relationships among the variables is as relevant as the specific way it is utilized within the evolutionary algorithm.

 \begin{table*}[htbp]
\begin{center}
\scriptsize
\begin{tabular}{|c|c|c|c|c|c|c|c|c|}
\hline\hline
   Algorithm & year  & NN model & Ref.   & Analysis Str. & Generative & Deep \\ \hline \hline  
    BEA      &2000 & HM   & \cite{Zhang_and_Shin:2000}            & no    &  yes  & no   \\ \hline 
    MONEDA   &2008 & GNG   & \cite{Marti_et_al:2008}              & no    &  no  & no   \\ \hline 
    RBM-EDA  &2010 & RBM  &  \cite{Tang_et_al:2010}               & no    &  no   & no  \\  \hline 
    REDA     &2013 & RBM  &  \cite{Shim_et_al:2013}               & no    &  yes  & no   \\ \hline 
    DAGA     &2014 & DA   &  \cite{Churchill_et_al:2014}          & yes   &  no   & no \\        \hline     
    DBM-EDA  &2015 & DBM  &  \cite{Probst_and_Rothlauf:2015}      & yes   &  yes  & yes  \\ \hline 
    AED-EDA  &2015 & DA   &  \cite{Probst:2015a}                  & yes   &  no   & no  \\ \hline 
    GAN-EDA  &2015 & GAN  &  \cite{Probst:2015}                   & no    &  yes  & no \\ \hline 
    GA-dA    &2016 & DA   &  \cite{Churchill_et_al:2016}          & yes   &  no   & no  \\ \hline 
    GA-NADE  &2016 & NADE &  \cite{Churchill_et_al:2016}          & yes   &  yes  & no \\ \hline 
    RBM-EDA &2017 & RBM   &  \cite{Probst_et_al:2017}      & yes   &  yes  & no  \\ \hline 
    Deep-Opt-GA  &2017 & DNN   & \cite{Baluja:2017}               &  yes  &  yes  & yes \\ \hline  
\end{tabular}
\caption{Description of some of the main  neural network models proposed for EDAs. Algorithms are organized following the chronological order of the publications. NN model refers to the neural model. HM: Helmzholtz machine; GNG: Growing Neural Gas; RBM: Restricted Boltzmann machine; DA: Denoising autoencoders; DBM: Deep Boltzmann machine; GAN: Generative adversarial network; NADE: Neural autoregressive distribution estimation; DNN: Deep NN. The rest of columns respectively indicate: whether the ``structure'' of the learned networks is inspected and discussed, whether the NN model is generative, whether the neural model used for the experiments is deep (contains $2$ or more hidden layers).}
\label{tab:NN_MODELS}
\end{center}
\end{table*}



One crucial dilemma in EDAs that learn complex graphical models is how to trade-off the pros and cons of the different classes of models. Bayesian networks are expensive to learn but can be efficiently sampled. Markov networks and factor graphs can be learned efficiently but usually require expensive iterative inference schemes (e.g., Gibbs sampling) to sample new solutions \cite{Muhlenbein:2012}.  In order for a neural model to be competitive with state-of-the-art graphical models  used in EDAs, they should be able to capture intricate relationships and, in addition, their  learning and sampling algorithms should outperform in efficiency those for graphical models. 

 The neural models that have been tried for EDAs, exhibit a variety of behaviors: autoencoders  used in a traditional EDA scheme  in \cite{Probst:2015a} are extremely fast when compared with methods that learn Bayesian networks but they fail to achieve the same efficiency than BOA in terms of function evaluations.  When used as a mutation distribution in \cite{Churchill_et_al:2016} GA-dA  (see Table~\ref{tab:NN_MODELS}) outperforms BOA in some problems (notably on the knapsack problem) but it is outperformed on the hierarchical HIFF function. As previously mentioned, for the performance of the EDA the class of model used  might be as relevant as the particular way it is used. Some of the neural models that have been tried, such as the generative adversarial network (GAN),  do not produce competitive results, neither in the number of fitness evaluations nor in the computational time \cite{Probst:2015}. 

The convenience of using deep neural models is another important question to discern given the impressive results of DNNs in other domains.  One of the conclusions obtained from the evaluation of the deep Boltzmann machine (DBM) neural network in EDAs  is that the effort for learning the multi-layered DBM model does not seem to pay off for the optimization process  \cite{Probst_and_Rothlauf:2015}. Also in \cite{Baluja:2017}, where DNNs with $5$ and $10$ layers are used as neural models, it is acknowledged that the learning process can be time consuming. While the Deep-Opt-GA is evaluated across a set of diverse artificial and real-world problems, it is not possible to determine the gain of the algorithm over EDAs since it is compared to a fast local optimizer and a GA.

In the following, we summarize a number of advantages and disadvantages of the use of neural models in EDAs.

Advantages of approaches that use NN models:

\begin{itemize}
 \item Flexible models. The type of dependencies of the neural model do not have to be specified a priori \cite{Baluja:2017}. 
 \item Usually can be efficiently (in terms of time) learned \cite{Churchill_et_al:2014,Probst:2015a}.  
 \item The gains in efficiency by means of parallelism (e.g., by using GPU architectures) can be dramatic \cite{Churchill_et_al:2016}. 
 \item They can be naturally applied to implement transfer learning approaches in EAs \cite{Churchill_et_al:2014}.
\end{itemize}

Disadvantages of approaches that use NN models:

\begin{itemize}
 \item Depending on the model, sampling can be cumbersome and costly. 
 \item NN models can be very sensitive to the initial parameters used for training. 
 \item The NN model representation does not always correspond to the representation of the problem (e.g., RBMs use binary representation, autoencoders use continuous representation, etc.). 
 \item The number of parameters required by the NN model can be very large (e.g., up to $2.5 \times n^2$ for autoencoders in some problems \cite{Churchill_et_al:2014}). 
 \item Regarding the number of parameters, overfitting can arise while learning the models \cite{Probst:2015a}. 
\end{itemize}

\subsection{Redefining problem structure for constrained and multi-objective problems}

 Two scenarios that present significant challenges for using the problem structure are multi-objective and constrained optimization problems.

 Addressing multi-objective problems (MOPs) is one of the most difficult areas for the implementation of methods that exploit the problem structure. The main difficulty is due to the variety and complexity of the modeling scenarios. One primary question is: What ``structure'' is the most relevant for MOPs? The structure describing the relationships among non-dominated points, or the (grouped or consensual) structure of all the objective functions involved ? Even if we assume that both types of structures are important, the question of finding a usable common representation for all the objectives is itself a challenge. The variety of selection, replacement, and archiving operators used by MOEAs determines important effects in the relationship  between the original problem structure and the interactions which are significant for the search. Finding a common ground for the investigation of the structure in this context is a very difficult task. 

 In EDAs, the question of how to take advantage of the problem structure for MOPs has been less explored than for single-objective optimization problems. Although several multi-objective EDAs (MO-EDAs) exist \cite{Bosman_and_Thierens:2002,Costa_and_Minisci:2003,Laumanns_and_Ocenasek:2002,Marti_et_al:2010,Pelikan_et_al:2006a,Zhang_et_al:2008,Zhou_et_al:2009} they have been proposed mainly for problems with a continuous representation. The characteristics of the search space for these problems allow other types of approaches which are different to those used for discrete problems. 

 In terms of the representation of the problem structure for MOPs, one idea proposed in \cite{Santana_et_al:2009b} was  \emph{the extension of the  representational capabilities of the probabilistic model to include information about the objectives}.  In  \cite{Karshenas_et_al:2014},  objectives were included as  additional nodes of the EDA probabilistic graphical model, and edges/arcs  were learned from data  to capture the relationships among variables and objectives. This initial work was successful when tested on continuous MOPs. Another work in the same direction, but addressing the combinatorial multi-objective knapsack problem, was presented in \cite{Martins_et_al:2016}. In the proposed approach, the graphical model represents, in addition to the decision variables and objectives, the parameters of a local optimization method. Other recent work \cite{Zangari_et_al:2016,Zangari_et_al:2015} has investigated  the structures of probabilistic models learned in different subproblems of the MOEA/D  algorithm \cite{Zhang_and_Li:2007}. Although this work has shown that the models in MOEA/D capture the structure of the problem, the convenience of using these structures to create factorizations for the  MOPs addressed in the paper was not addressed. 

 In \cite{Chicano_et_al:2016a}, a multi-objective Hamming-ball hill climber is proposed for a MOP, where each objective is a  pseudo-Boolean functions with k-bounded epistasis. This gray-box optimization algorithm uses as a representation of the problem structure  a co-occurrence graph in which two variables are joined by an edge if they co-occur in any of the subfunctions for any of pseudo-Boolean functions. The introduced algorithm guarantees that  each move is bounded in constant time if each variable appears in a constant  number of subfunctions. The algorithm is tested on instances of the MNK-Landscape with two and three objectives and up to $100,000$ variables. Although it is impossible to determine how close the results are to the optimal Pareto front, it shows that they improve by increasing the radius of the hill climber, i.e., considering higher order dependencies among the variables.

  The difficulty with the search in a constrained space is that the original interactions of the problems can be distorted due to the constraints. Similarly, constraints can produce new interactions among the variables. Early work investigated how to modify FDAs to deal with problems with unitation constraints \cite{Santana_and_Ochoa:1999b,Santana_et_al:2001c}. This research showed that while constraints can have an impact on  the strength of dependencies among those variables related by the problem structure, it was still possible to exploit the original structure of the problem to make a more efficient search. Recent approaches propose the use of probability models defined exclusively on the space of feasible solutions \cite{Ceberio_et_al:2017}.
 
 As expected, constraints also have an effect on the behavior of gray-box optimizers. In \cite{Chicano_et_al:2016}, it is shown that the constant time per move of gray-box hill climbers is not guaranteed to be achieved when constraints are added to the problem. However, it is also shown in \cite{Chicano_et_al:2016} that the proposed structure-informed hill-climber has as worst-case run-time $O(n)$ for constrained multi-objective $k$-bounded pseudo-Boolean problems.
 

\subsection{The question of evolvability}

 Evolvability has been defined in numerous ways, and the implications of the term both in the biological and evolutionary computation domains are controversial. It can be defined as an organism's capacity to generate heritable phenotypic variation \cite{Kirschner_and_Gerhart:1998}, the  increased potential of an individual or population to further evolution \cite{Mengistu_et_al:2016}, or the ability of random variations to sometimes produce improvement \cite{Wagner_et_al:1996}. Recently,  Wilder and Stanley  \cite{Wilder_and_Stanley:2015} have advocated for making a difference between the concepts \emph{evolvable individuals} and \emph{evolvable populations}. Evolvable individuals are more likely than others to introduce phenotypic variation in their offspring, and in evolvable populations a greater amount of phenotypic variation is accessible to the population as a whole, regardless of how evolvable any individual may be in isolation  \cite{Wilder_and_Stanley:2015}.

From the perspective of the analysis developed in this paper, one key question is whether the use of the problem structure information in EAs hinders or promotes the evolvability of solutions and populations. The focus of gray-box optimizers, EDAs, and other algorithms that exploit the problem structure has been  efficient function optimization, with the number of function evaluations as the main target. However, it is not clear whether designing model-based operators and algorithms that constrain the way genotypic variation is manifested through evolution can also decrease the capacity of these evolutionary systems to adapt to changes. 

Much of function benchmarks on which the algorithms discussed in this paper have been tested exclusively comprise non-dynamic objective functions with a simple genotype-phenotype mapping. Consequently,  it is difficult to evaluate how robust these algorithms would be to deal with dynamic  functions, particularly with those functions where the structure of interactions changes over time. It can be argued that model-based EAs that learn the model from the data have a natural way to adapt to potential changes in the fitness function. The ability of gray-box optimizers to treat these problems is not clear. Whatever the capacity of these algorithms to deal with dynamic functions is, what remains as an open question is whether the learning and sampling methods used by model-based EAs constrain, promote, or are neutral in generating evolvable individuals or populations. 

  There is an  intense debate around the question of whether and how evolvability evolves \cite{Brookfield:2001,Earl_and_Deem:2004,Kirschner_and_Gerhart:1998,Radman_et_al:1999}. Related questions are:  how genes create phenotypes in such a way that options for future evolution are enhanced \cite{Brookfield:2001}?; is evolvability a consequence of adaptation and/or selection \cite{Earl_and_Deem:2004,Kirschner_and_Gerhart:1998,Sniegowski_and_Murphy:2006}?; is evolvability influenced by the existence of neutral networks in genotype space \cite{Ebner_et_al:2001}? 

   In the context of model-based EAs one relevant topic is to what extent can the evolution of evolvability be modeled and eventually biased by exploiting the model.  If evolvability can be evolved, is it possible to learn and exploit the structure of systems that evolve evolvability? Which models could be used to learn the patterns behind the individuals propensity to evolve? Another related question is whether problem structure plays any role in the algorithms that try to directly evolve for evolvability or in those EAs which indirectly encourage evolvability  \cite{Kashtan_and_Alon:2005,Lehman_and_Miikkulainen:2015,Lehman_and_Stanley:2011,Mengistu_et_al:2016}.

\subsection{Combination of gray-box optimization and EDAs}

A straightforward way of combining gray-box optimization and EDAs is to use gray-box optimization as a local search technique within EDAs. This is the approach followed in \cite{Goldman_et_al:2015}, where the HBHC is combined with the Parameter-less population pyramid (P3) \cite{Goldman_and_Punch:2014}, a model-building EA. In fact, when successful for real-world and large optimization problems, the vast majority of EDAs require the application of local optimizers \cite{Muehlenbein_and_Mahnig:2002,Pelikan:2005,Santana_et_al:2008f,Zhang_et_alL2003a}. One of the conclusions from the research presented in  \cite{Goldman_et_al:2015} is that gray-box optimization by itself may fail to obtain optimal results and the hypothesis that this may be due to the existence of plateaus is advanced. A simple hybrid algorithm combining uniform crossover with HBHC is able to outperform the HBHC. 

 Several works  in EAs show the great success of a variety of local optimization methods \cite{Pelikan_et_al:2003a,Radeti_et_al:2009,Santana_et_al:2008f}  when combined with EDAs. Similarly, the emergence of new paradigms such as Optimal Mixing indicates that the possibilities to combine local search and  problem structure have not been exhausted yet. Therefore, it will require an in-depth investigation to determine to what extent  gray-box local search optimizers can outperform the results of the other hybrid approaches.

\section{Conclusions} \label{sec:CONCLU}

 While pioneering work in EAs showed the importance of exploiting problem information, work in FDAs, gray-box optimization, optimal mixing, and other approaches reveals that, when information about the problem structure is available in the form of a graphical representation, a variety of methods can be designed to efficiently exploit this structure. The different ways in which these methods exploit the structure reveal the great, largely unexplored, potential of approaches that use graphical models.

 In this paper we have reviewed algorithms that exploit the problem structure in evolutionary computation with a focus on gray-box optimizers and FDAs.  We have used  a common perspective to analyze these algorithms, highlighting the similarities and differences among them.  We have contrasted the explanations of the EA behavior based on the schema analysis with those that emphasize the role of factors and factorizations, and presented examples of these characterizations at work. Optimal mixing algorithms, as a hybrid approach that combines characteristics from EDAs and GAs have been also covered. 

 The White-Gray-Black (WGB) classification of optimization problems, according to the type and extent of problem information available, has been introduced as a more refined way to identify the use of problem structure. A review of recently introduced neural approaches used as models has been conducted. Finally, the paper has addressed a number of open questions and hot-topics where opportunities for algorithms that exploit the problem structure arise.

\section*{Acknowledgment}

This work has received support from the IT-609-13 program (Basque Government),  TIN2016-78365-R (Spanish Ministry of Economy, Industry and Competitiveness), and  CNPq Program Science Without Borders Nos.: 400125/2014-5 (Brazil Government)


\end{document}